\documentclass[10pt,twocolumn,letterpaper]{article}
\usepackage{titling}
\usepackage[pagenumbers]{cvpr} 

\usepackage[accsupp]{axessibility}
\usepackage{graphicx}
\usepackage{amsmath}
\usepackage{amssymb}
\usepackage{booktabs}
\usepackage{xcolor, color, colortbl}
\usepackage{multirow}
\usepackage{pifont}
\usepackage{array}

\usepackage[pagebackref,breaklinks,colorlinks]{hyperref}

\usepackage[capitalize]{cleveref}
\crefname{section}{Sec.}{Secs.}
\Crefname{section}{Section}{Sections}
\Crefname{table}{Table}{Tables}
\crefname{table}{Tab.}{Tabs.}

\newcommand{\cmark}{\ding{51}}%
\newcommand{\xmark}{\ding{55}}%
\newcommand{\supp}[1]{\color{magenta} #1 \color{black}}
\newcommand{\net}{TemporalStereo}
\newcommand{\nonlocal}{Statistical Fusion }

\definecolor{ours}{RGB}{1,1,0.7}
\definecolor{w1}{RGB}{153,255,153}
\definecolor{w4}{RGB}{153,204,255}
\definecolor{w8}{RGB}{255,153,153}
\definecolor{lightwhite}{RGB}{255,255,255}
\definecolor{lightblue}{RGB}{10,255,255}

\begin{document}

\title{TemporalStereo: Efficient Spatial-Temporal Stereo Matching Network}

\author{Youmin Zhang \hspace{0.5cm} Matteo Poggi \hspace{0.5cm} Stefano Mattoccia \\
CVLAB, Department of Computer Science and Engineering (DISI) \\
University of Bologna, Italy\\
\normalsize{\url{https://youmi-zym.github.io/projects/TemporalStereo/}} \\
\texttt{\{youmin.zhang2, m.poggi, stefano.mattoccia\}@unibo.it}
}

\maketitle

\begin{abstract}

We present TemporalStereo, a coarse-to-fine stereo matching network that is highly efficient, and able to effectively exploit the past geometry and context information to boost matching accuracy. Our network leverages sparse cost volume and proves to be effective when a single stereo pair is given. However, its peculiar ability to use spatio-temporal information across stereo sequences allows TemporalStereo to alleviate problems such as occlusions and reflective regions while enjoying high efficiency also in this latter case. Notably, our model -- trained once with stereo videos -- can run in both single-pair and temporal modes seamlessly. Experiments show that our network relying on camera motion is robust even to dynamic objects when running on videos. We validate TemporalStereo through extensive experiments on synthetic (SceneFlow, TartanAir) and real (KITTI 2012, KITTI 2015) datasets. Our model achieves state-of-the-art performance on any of these datasets. 

\end{abstract}

\begin{figure}[t]
    \centering
    \begin{tabular}{c}
    \includegraphics[width=0.45\textwidth]{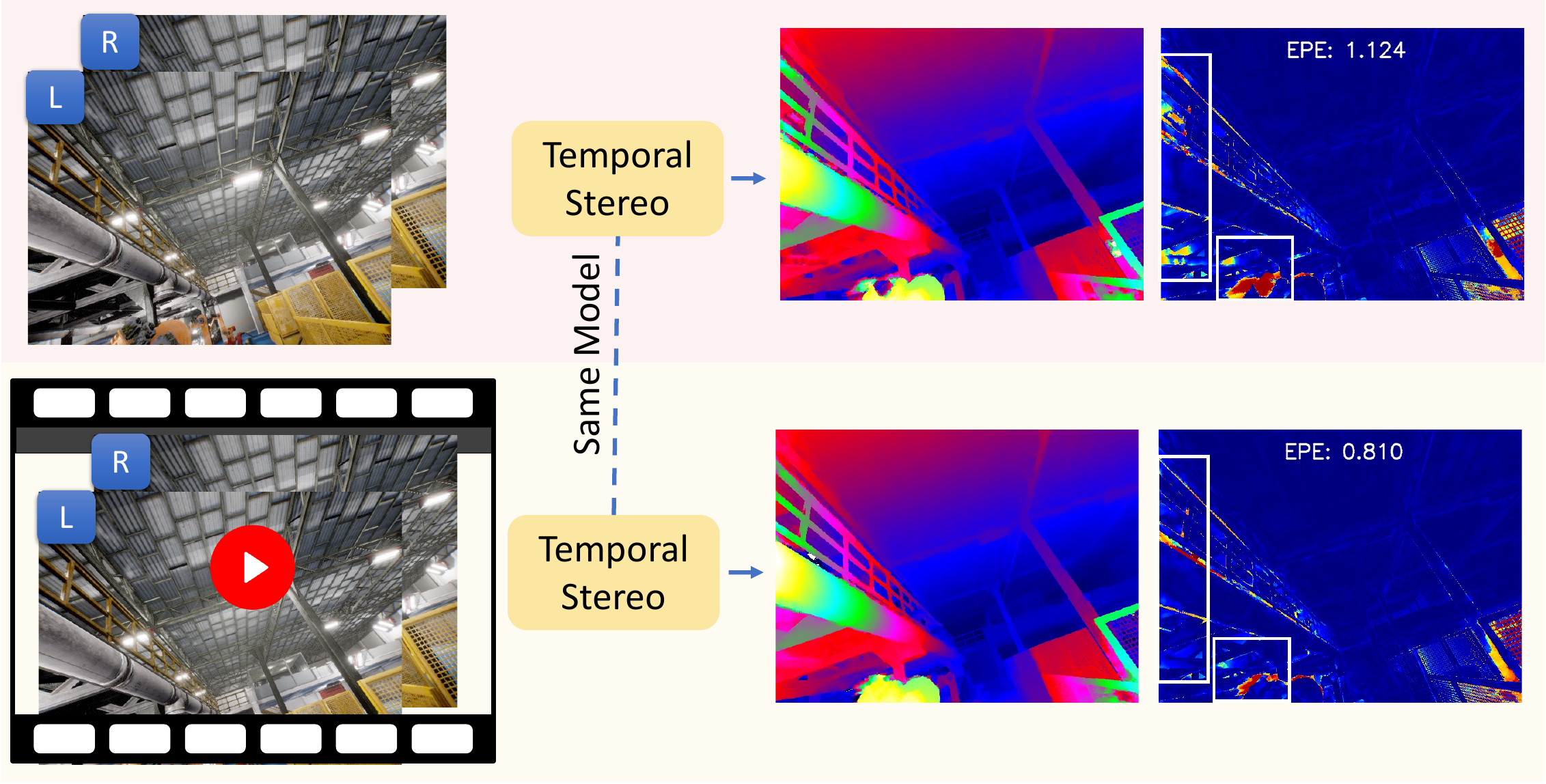} \\
    \includegraphics[width=0.45\textwidth]{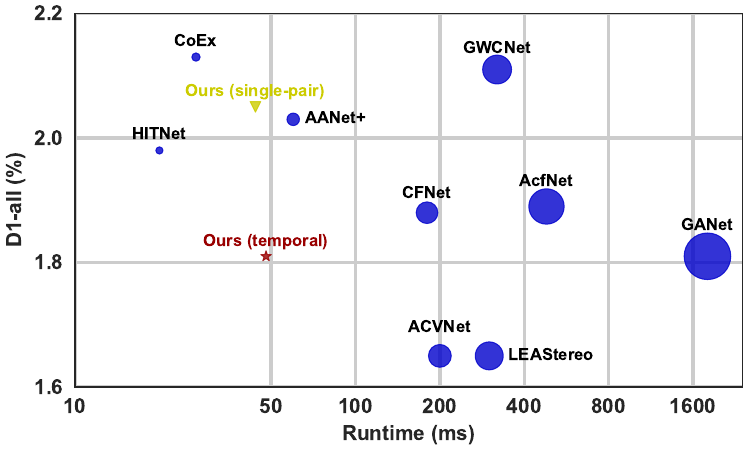} \\
    \end{tabular}
    \caption{\textbf{Overview of TemporalStereo.}
    Our model can seamlessly process a stereo pair, as well as switch to temporal mode when multiple pairs are available, to increase accuracy. This allows TemporalStereo for achieving a favorable accuracy/speed trade-off with respect to existing networks. GMacs are measured with a resolution of 960×572.}
\label{fig:arch_performance}
\end{figure}

\section{Introduction}

Stereo reconstruction is a fundamental problem in computer vision. It aims at recovering the 3D geometry of the sensed scene by computing the disparity between corresponding pixels in a rectified pair of images. In robotic~\cite{robot}, virtual/augmented reality~\cite{augmented-reality}, autonomous driving~\cite{auto-driving}, and more, depth information is crucial for scene understanding, with accurate yet fast solutions being particularly attractive.

Recently, deep learning approaches have shown tremendous improvements, especially in the supervised setting~\cite{ACVNet,LEAStereo,CREStereo}. Common to most architectures is the cost volume, which encodes the feature similarity and plays a key role in matching pixels. In particular, features from the left image are concatenated (in 3D networks) or correlated (in 2D networks) and processed through convolutions, which are costly and prevent real-time execution. To address this issue, more efficient stereo models have been proposed, taking advantage of efficient cost aggregation~\cite{coex,AANet}, coarse-to-fine~\cite{StereoNet,AnyNet} and cost volume pruning~\cite{DeepPruner,UASNet,CFNet} strategies to preserve computation and accuracy. Nonetheless, the sparse and constant disparity candidates used for matching cost computation can easily miss the ground truth disparity, thus deteriorating the accuracy of prediction. The case becomes even worse when dealing with challenging regions. For example, although occlusion is an ill-posed problem for stereo matching based on a single stereo pair, popular supervised stereo methods \textit{have not ever} taken advantage of the intrinsic correlation among frames when stereo videos are available. Even if commercial cameras can easily acquire high FPS videos (e.g., 30 or more), in which likely past context is strictly related to the current one, state-of-the-art stereo methods generally process each frame independently.  We argue that previous disparities can be relevant cues to improve estimates, especially in complex regions where the current pair alone is not sufficiently meaningful, such as near object boundaries and occlusions.

Inspired by these observations, we propose \textit{TemporalStereo}, a novel lightweight deep stereo model (as depicted in Fig. \ref{fig:arch_performance}) being not only suited for single pair fast inference, but also able to leverage the rich spatial-temporal cues provided by stereo videos, when available. In \textit{single-pair} mode (i.e., when only the current pair is available), our network follows a coarse-to-fine scheme~\cite{AnyNet,StereoNet,CasStereo} to enable efficient inference. 
In order to increase the capacity of the constructed sparse cost volume and also the quality of sparse disparity candidates at every single stage, we enhance the cascade-based architecture in several aspects: 1) firstly, we enrich the cost of each queried disparity candidate with context from non-local neighborhoods during cost computation; 2) as the number of disparity candidates could be decreased to very few (e.g., 5), in addition to exploiting pure 3D convolutions for cost aggregation, we also perform multiple statistics (e.g., mean and maximum statistics) over a large window to grasp the global cost distribution of each pixel at one stage; 3) differently from previous strategies~\cite{UASNet,StereoNet,AnyNet,CFNet}, for which the candidates are constant throughout each stage, we rely on the aggregated cost volume itself to \textit{shift} the candidates towards a better location and thus improve the quality of predicted disparity map; 4) additionally, when multiple pairs are available, our network can easily switch to \textit{temporal} mode, in which past disparities, costs and cached features could also be aligned to the current reference frame and used to boost current estimates with a negligible runtime increase (4ms as shown in Tab.~\ref{tab:benchmark}). Experimental results on synthetic (SceneFlow, TartanAir) and real (KITTI 2012, KITTI 2015) benchmarks support the following claims:

\begin{itemize}
    \item Our method is accurate yet fast and achieves state-of-the-art results when a single stereo pair is available. In particular, the improved coarse-to-fine design based on very few (e.g., 5) candidates allows \net{} to unlock fast execution and high performance.
      
    \item When multiple temporally adjacent stereo pairs are available, \net{} is the first supervised stereo network that can effortlessly cache the past context and use it to improve ongoing predictions, e.g., in occluded and reflective regions. The model trained in temporal mode is effective also in single-pair mode, allowing the deployment of the same model in both settings. 
    
    \item The proposed temporal cues can be widely applied to boost the matching accuracy of current efficient stereo methods, e.g., on TartanAir~\cite{TartanAir} dataset, the improvements of CoEX~\cite{coex} and StereoNet~\cite{StereoNet} are 14.6\% and 26.1\% respectively. Nonetheless, our \net{} is still the best one for utilizing temporal information. 
\end{itemize}

\begin{figure*}[!tbp]
	\centering
		\includegraphics[width=2.0\columnwidth]{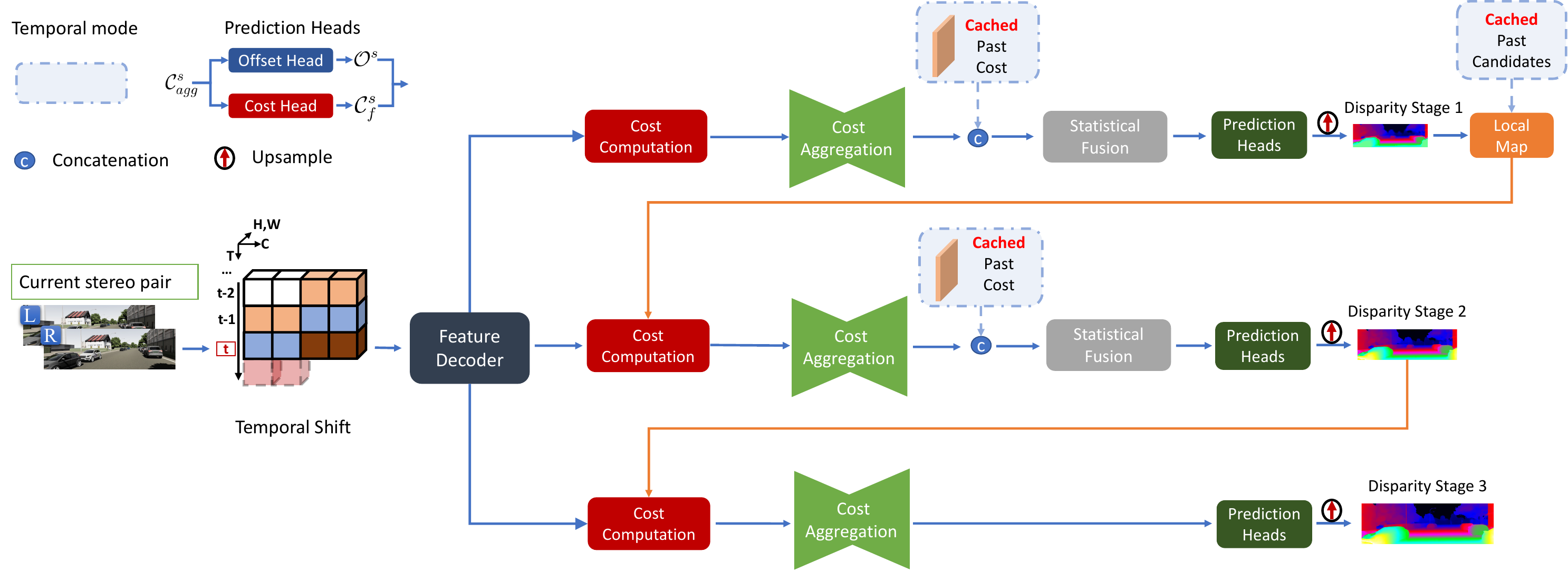}
		\caption{\textbf{\net{} Architecture}. In single-pair mode, the model predicts the disparity map in a coarse-to-fine manner.
		If past pairs are available, the same model switches to temporal mode and employs features, costs, and candidates cached from past pairs to improve the current prediction. 
		} 
	\label{fig:framework}
\end{figure*}

\section{Related Work}

\textbf{Single Pair Stereo Matching.} Traditional methods \cite{fastfilter} employ handcrafted schemes to find local correspondences~\cite{4steps,nonlocalaggregation} and approximate global optimization by exploiting spatial context~\cite{SGM}. Recent deep learning strategies \cite{poggi2021synergies} adopt CNNs and can be broadly divided into two categories according to how they build the cost volume. On the one hand, 2D networks follow DispNetC~\cite{sceneflow} employing a correlation-based cost volume and applying 2D convolutions to regress a disparity map. Stacked refinement sub-networks \cite{iResNet} and multi-task learning~\cite{edgestereo,segstereo} have been proposed to improve the accuracy. GCNet~\cite{GCNet} represents a milestone for 3D networks. 
Following works improve its results thanks to spatial pyramid pooling \cite{PSMNet}, group-wise correlations~\cite{GWCNet}, forcing unimodal \cite{AcfNet,wasserstein} or bimodal \cite{SMDNet} cost distributions. 
3D methods usually outperform 2D architecture by a large margin on popular benchmarks~\cite{sceneflow,KITTI2012,KITTI2015}, paying more in terms of computational requirement. In this work, we leverage 3D sparse cost volumes, and the peculiar proposed architecture proves to be fast and accurate at disparity estimation.

\textbf{Efficient Stereo Matching with Deep-Learning.} 
A popular strategy to decrease the runtime consists in performing computation at a single yet small resolution (e.g., 1/4) and obtaining the final disparity through upsampling. CoEx~\cite{coex} adopts this strategy. Coarse-to-fine \cite{BI3D,HSM} design 
further improves efficiency, since the overall computation is split into many stages. To further reduce the disparity search range for each pixel, StereoNet~\cite{StereoNet} and AnyNet~\cite{AnyNet} add a constant offset to disparity maps produced in the previous stage to avoid checking all the disparity candidates. However, as reported in \cite{UASNet}, the constant offset strategy is not robust against large errors in initial disparity maps. DeepPruner \cite{DeepPruner} prunes the search space using a differentiable variant of PatchMatch \cite{barnes2009patchmatch}, obtaining sparse cost volumes. In contrast, methods such as \cite{UASNet,CFNet} employ uncertainty to evaluate the outcome of previous stages and then build the current cost volume accordingly. However, since the cost volume itself expresses the goodness of candidates (the better the candidates, the more representative the cost volume), it could be used to \textit{check} and eventually \textit{correct} the candidates themselves. Nonetheless, previous methods in literature cannot exploit this cue since their candidates come from the earlier level and are constant through the current stage. Conversely, this paper proposes inferring an offset for each disparity candidate from the current aggregated cost volume, which helps to ameliorate the disparity. Moreover, to improve network efficiency on high-resolution images, we perform the coarse-to-fine predictions only at downsampled feature maps rather than at full image resolution like HITNet~\cite{HitNet}.

\textbf{Multiple Pairs Stereo Matching.}
When two temporally adjacent stereo pairs are available, optical flow~\cite{sceneflow,raft} or 3D scene flow~\cite{teed2021raft3d} is often taken into account to link the images with the reconstructed 2D/3D motion field. Then, the stereo problem is tackled by leveraging a multi-task model \cite{aleotti2020learning,Lai2019bridging,unos} or by casting it as a special case of optical flow \cite{flow2stereo}. However, these methods are not able to capitalize longer stereo sequences. Conversely, OpenStereoNet \cite{zhong2018open} adopts recurrent units to capture temporal dynamics and correlations available in videos for unsupervised disparity estimation. We argue that past information could be beneficial to understand the current scene -- especially in difficult areas -- for \textit{free} since past context and predictions can be cached easily, although none of the above methods make use of them explicitly in videos by taking into account geometry. On the contrary, \net{} fully exploits this potential.

\section{Method}

\begin{figure*}[!t]
    \centering
    \includegraphics[width=1.8\columnwidth]{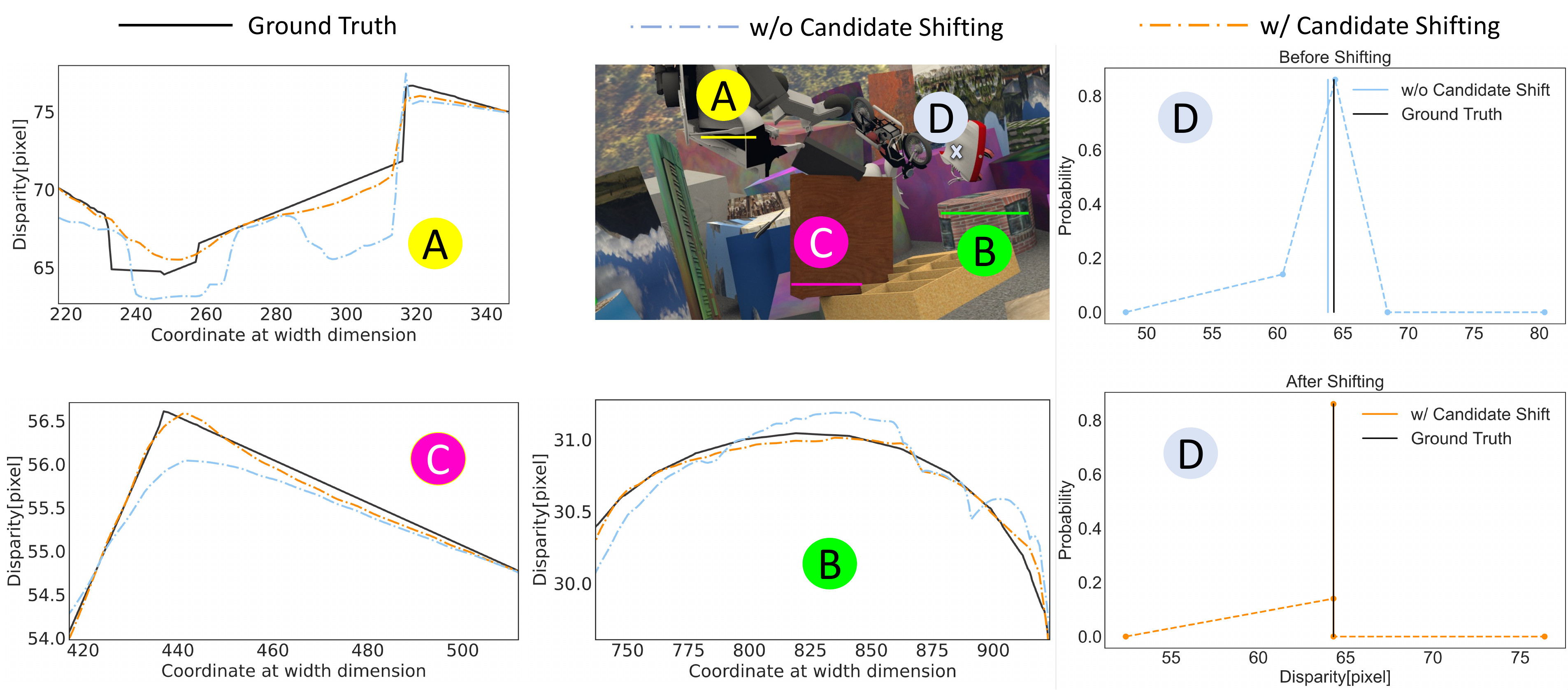}
    \vspace{-0.1cm}
    \caption{\textbf{Adaptive Shifting in action.} We show the ground truth disparities and the predictions with and without adaptive shifting for three horizontal regions(i.e., A, B, C) and one pixel (D) in the image. For A, B, C, our strategy better estimates the actual disparity distributions.  
    For D, the matching distribution of 5 disparity candidates switches from a flattened unimodal to one-hot distribution centered at ground truth by adaptively shifting candidates.} 
    \label{fig:adaptive-shifting}
\end{figure*}

We first describe \net{} in single-pair mode,
then we fully unlock the potential of our architecture enabling temporal mode. Fig. \ref{fig:framework} depicts our model.

\subsection{Single Pair Mode}
Given a single stereo pair, a backbone extracts multi-scale features at $1/4$, $1/8$, and $1/16$ of the original resolution. Then, three stages predict disparity maps starting from these features. In particular, each stage performs feature decoding, cost volume computation, and aggregation. 

\textbf{Feature decoding}
In each stage $s\in\{1,2,3\}$, a decoder processes the current features, together with those from the previous stage $F_{l}^{s-1}$, $F_{r}^{s-1}$ if $s>1$. In particular, $F_{l}^{s-1}$, $F_{r}^{s-1}$ are bilinearly upsampled by a factor 2 and concatenated with left and right features from the backbone. Then, the resulting feature maps are processed by two 2D convolutions with kernels of 3, obtaining $F_{l}^{s}$, $F_{r}^{s}$.

\textbf{Cost volume computation.}\label{cost-computation} A 4D cost volume $\mathcal{C}^{s}\in \mathbb{R}^{Ch^s\times H^s\times W^s\times |\mathcal{D}^s| }$ is constructed by concatenating $F_{l}^{s}$ with the corresponding $F_{r}^{s}$ for each disparity $d$ in the set $\mathcal{D}^s=\{d_1, d_2, ..., d_{n}\}$, with $Ch^{s}$ respectively the number of channels of feature $F_{l}^{s}$ and disparity candidates used in the stage:
\begin{equation}
    \begin{aligned}
    \mathcal{C}_{concat}^{s}(\cdot, u, v, d) = \oplus \{F_{l}^{s}(\cdot,u,v), F_{r}^{s}(\cdot, u-d,v)\},
    \end{aligned}
    \vspace{-0.1cm}
    \label{eq:concat}
\end{equation}
where $u,v$ are horizontal and vertical pixel coordinates, respectively, while $\oplus$ stands for concatenation on channel dimension. However, by doing so, $\mathcal{C}^s$ only contains local information, which is a major limitation in the case of a sparse set of candidates. To alleviate it, we enrich $\mathcal{C}^{s}$ with multi-level costs~\cite{raft} encoding a larger neighborhood. Specifically, we build a three-level cost volume from feature maps $F_{l}^{s},\; F_{r}^{s}$ and by downsampling them with factors $2,4$. At each level, we perform group-wise correlations \cite{GWCNet}, then costs are bilinearly upsampled to $\mathcal{C}^s$ resolution and stacked together, obtaining $\mathcal{C}_{gwc}^s$. The final multi-level cost is:
\begin{equation}
    \begin{aligned}
        \mathcal{C}^{s}(\cdot, u, v, d) = \oplus \{ \mathcal{C}_{concat}^{s}(\cdot, u, v, d),  \mathcal{C}_{gwc}^{s}(\cdot, u, v, d) \}.
    \end{aligned}
    \vspace{-0.1cm}
\end{equation}

\textbf{Cost aggregation.}
We now describe how we compute the aggregated cost volume $\mathcal{C}^{s}_{agg}$. In each stage, preliminary spatial cost aggregation is performed by one 3D convolution with kernel size $3\times3\times3$ followed by an hourglass network and another 3D convolution with kernel size $3\times3\times3$. Afterward, a \nonlocal module further improves the aggregation with mean and maximum statistics~\cite{SepFlow}, especially along the disparity dimension (which contains as few as 5 candidates): 4 parallel layers (identity, convolutional with kernel size $5\times1\times1$, average pooling and max pooling both with kernel size $5\times5\times5$, respectively) extract global statistics from the preliminary aggregated volume, then their outcomes are stacked together and processed by one 3D convolution.
In temporal mode, \nonlocal also helps to merge past costs with current ones, as we will discuss later. We employ \nonlocal only in stages $s=1,2$ to lighten the computation, while in $s=3$ past costs are not used. Finally, a cost prediction head with two 3D convolutions with kernel size $3\times3\times3$ is in charge of predicting the final cost volume $\mathbf{C}^{s}_{f}$ from $\mathcal{C}^{s}_{agg}$. From $\mathcal{C}^{s}_{f}$ and $\mathcal{D}^s$ we could obtain the disparity $\hat{d}^{s}$ of the current stage by means of soft-argmin operator \cite{GCNet}, but $\mathcal{C}^{s}_{agg}$ can help us to improve the candidates -- as we will describe thereafter.
For this reason, before computing $\hat{d}^{s}$, we have to introduce our adaptive candidate shifting strategy.

\textbf{Adaptive candidate shifting.} Previous works \cite{UASNet,StereoNet} consider the candidates $\mathcal{D}^s$ as constants throughout the current stage $s$. However, the aggregated cost volume itself contains cues about the candidates used to build it. Since it has been constructed from $\mathcal{D}^s$, it represents how well the candidates can explain the current scene. Moreover, in the cost aggregation step, each cost has been enriched with all the costs in its neighborhood, collecting matching results of nearby pixels. We propose to leverage a dedicated prediction head to infer, from $\mathcal{C}^{s}_{agg}$ as shown in Fig. \ref{fig:framework}, an offset value for each candidate, which represents how much we have to \textit{shift} the candidate towards a better location. Thus, the candidates are pixel-wisely different and  able to adaptively adjust according to the learned context. This strategy gives the network the ability to adapt $\mathcal{D}^s$ and improve $\hat{d}^s$, as depicted in Fig. \ref{fig:adaptive-shifting}. Specifically, an independent head -- with the same structure of the cost prediction head, i.e., two 3D convolutions, -- is in charge of inferring the offset volume $\mathcal{O}^{s}\in \mathbb{R}^{H^s\times W^s\times |\mathcal{D}^s|}$ from $\mathcal{C}^{s}_{agg}$ as input. Although offset learning~\cite{liu2016ssd} is not new and has been used in~\cite{wasserstein} to convert the dense disparity candidates, which always tightly cover the ground truth disparity, into continuous ones so that the predicted disparity value will not be restricted to integers, in contrast, our disparity candidates are few and sparse, which could easily miss the ground truth. By enriching the context of each sparse candidate, the proposed shifting strategy is able to \textit{correct} the constant and flawed candidates for better disparity estimation.

\textbf{Disparity prediction and candidate sampling.}
Given $\mathcal{D}^s$, $\mathcal{O}^{s}$ and $\mathcal{C}^{s}_{f}$, soft-argmin operator~\cite{GCNet} could be applied to predict the final disparity map. However, this strategy results sub-optimal in the case of multi-modal distributions~\cite{AcfNet,wasserstein,SMDNet}. To overcome this limitation, we predict the disparity following the strategy proposed in~\cite{coex}: we select top-$K$ values of $-\mathcal{C}^{s}_{f}$ for each pixel across the disparity dimension, and we normalize them with the softmax operator $\sigma$. The predicted disparity map $\hat{d}^{s}$ results:
\begin{equation}
    \begin{aligned}
    \hat{d}^{s} =  \sum_{d\in\{d^{top}_{1}, \cdots, d^{top}_{K}\}}  (d + o_{d}) \times \sigma(-c_{d}),
    \end{aligned}
    \vspace{-0.2cm}
    \label{eq:softargmin}
\end{equation}
where $d^{top}_{k} = {argmax}_d^{k}(-c_{d})$, $argmax^{k}(\cdot)$ is the $k$-th maximal value for $k\in\{1, 2, ..., K\}$, $c_{d}$ the cost in $\mathcal{C}^{s}_{f}$ of candidate $d$ for each pixel, and $o_{d}$ is the offset value in $\mathcal{O}^{s}$ for the candidate $d$. Convex upsampling \cite{raft} is used to double the resolution of $\hat{d}^1$ and $\hat{d}^2$, obtaining $\hat{d}^1_\uparrow$, $\hat{d}^2_\uparrow$. To bring $\hat{d}^3$ to full resolution, following~\cite{coex}, we bilinearly upsample $\hat{d}^3$ to full resolution and for each pixel in the upsampled resolution, a weighted average of a $3\times3$ superpixel surrounding it is calculated to get the final prediction $\hat{d}^3_\uparrow$. 
Finally, we obtain the set of candidates $\mathcal{D}^{s+1}$ for the next stage applying an inverse transform sampling. We sample according to a normal distribution $\mathcal{N}(\hat{d}^{s}_\uparrow, 1)$ in the range $[\hat{d}^{s}_\uparrow - \beta, \hat{d}^{s}_\uparrow + \beta]$, with $\beta$ a constant value. This strategy allows to include more candidates near the predicted disparity. The first stage is initialized with $\mathcal{D}^1$ uniformly sampled across the full range to provide an overview of current scene geometry.

\textbf{Loss function.} Following~\cite{PSMNet}, we minimise the smooth $L_1$ loss, i.e., Huber loss at full resolution between the ground truth disparity map $d^{gt}$ and the predictions at the three stages $\{\hat{d}^{1}_\uparrow, \hat{d}^{2}_\uparrow, \hat{d}^{3}, \hat{d}^{3}_\uparrow\}$. $\hat{d}^{1}_\uparrow$, $\hat{d}^{2}_\uparrow$ and $\hat{d}^{3}$ are bilinearly upsampled to full resolution and a factor $\lambda^s$ weights each predicted map. The loss results:
\begin{equation}
    \begin{aligned}
    \mathcal{L}_{huber} & =  \frac{1}{|\mathcal{N}|} \lambda^{0} \cdot smooth_{L_{1}}(d^{gt} - \hat{d}^{3}) \\ &+ \sum_{s =1}^{3} \frac{1}{|\mathcal{N}|} \lambda^{s} \cdot smooth_{L_{1}}(d^{gt} - \hat{d}^{s}_\uparrow),
    \end{aligned}
    \vspace{-0.1cm}
    \label{eq:l1loss}
\end{equation}
where $\mathcal{N}$ is the number of pixel with valid ground truth.

Furthermore, to supervise the learning of offsets in three stages $\{{O}^{1}, {O}^{2}, {O}^{3}\}$, we adopt the Warsserstein loss~\cite{wasserstein} for training. Specifically, for each stage $s$, we downsample the ground truth disparity map $d^{gt}$ to the resolution of corresponding offset for supervision, obtaining $d^{gt, s}_{\downarrow}$. As the offset value can range in a quite large value space, we improve the Warsserstein loss as:
\begin{equation}
    \begin{aligned}
    \mathcal{L}_{war} = \sum_{s =1}^{3}\frac{1}{|\mathcal{N}|} \lambda^{s} \cdot \sum_{d\in \mathcal{D}^{s}} \left(  |d + o_{d} - d^{gt,s}_{\downarrow}| \times \left(\sigma(-c_{d}) + \alpha\right) \right)
    \end{aligned}
    \label{eq:warloss}
\end{equation}
where $c_{d}$ the cost value in $\mathcal{C}^{s}_{f}$ of candidate $d$ for each pixel, and $o_{d}$ is the offset value in $\mathcal{O}^{s}$ for the candidate $d$. We set $\alpha = 0.25$, i.e., even for the candidate with extremely low matching probability $\sigma(-c_{d})$, the network still learns an offset to enforce it moving towards the ground truth. With a weight factor $\lambda_{final}$, our final loss  becomes:
\begin{equation}
    \begin{aligned}
    \mathcal{L} = \mathcal{L}_{huber} + \lambda_{final} \cdot \mathcal{L}_{war}.
    \end{aligned}
    \label{eq:allloss}
\end{equation}

\subsection{Temporal Mode}
So far, we have presented the model suited for single-pair mode. Now, we illustrate how the model works in temporal mode. Differently from multi-view stereo models \cite{deepmvs,deepv2d}, our network processes stereo videos one stereo pair at a time. This behaviour, which helps to save computation, is possible thanks to the sparse cost volume. In fact, we can easily add disparity candidates from the past inferences to the current set $\mathcal{D}^s$, thus increasing the search range with other plausible solutions. 
The past candidates, however, have to be aligned with the current frame to be meaningful. Optical flow/3D scene flow could be used to tackle the issue, but predicting flow fields is expensive in terms of memory footprint and time. Instead, in this work, we suppose that the camera is calibrated and the pose is given -- since pose can be provided by external IMU sensors or estimated from the stereo pairs-- to build the rigid motion field needed to align the candidates. Nonetheless, rigid flow formulation only holds for static objects in the scene, but since our network always relies on the current stereo pair, \net{} is robust even in case of wrong flows (e.g., due to bad poses) or moving objects. In the remainder, we detail how we do use temporal information to improve stereo-matching results.
 
\textbf{Local map.} \label{local-map} As highlighted before, occlusions represent a major issue in stereo matching. However, we argue that currently occluded regions might have been visible in past frames, e.g., due to camera motion. For this reason, the issue can be alleviated by adequately exploiting past estimates and at a minimal cost since they can be easily cached. Furthermore, we cache costs according to a keyframe strategy to bind the complexity in the case of long video sequences and ensure enough motion parallax. Following~\cite{neuralrecon}, a new incoming frame is promoted to keyframe if its relative translation and rotation are greater than $\textbf{t}_{max}$ and $\textbf{R}_{max}$ respectively. Then, a memory bank collects disparity $\hat{d}^3_\uparrow$ of the last $N_{key}$ keyframes. Since each cached map represents the scene geometry in the past, to use it in the current computation we need to update the disparity values and their coordinates in the current image. Inspired by SLAM literature \cite{murORB2}, we leverage a Local Map strategy to this aim. In detail, given the camera model $\pi:\mathbb{R}^{3} \rightarrow \mathbb{R}^{2}$, a 3D point $P(X, Y, Z)$ can be projected to a 2D pixel $p(u,v)$:
\begin{equation}
    \begin{aligned}
    \pi(P) = \left( f_{x}\frac{X}{Z}+c_{x}, f_{y}\frac{Y}{Z}+c_{y} \right),
    \end{aligned}
    \vspace{-0.1cm}
\end{equation}
where $(f_x, f_y, c_x, c_y)$ are the known camera intrinsics. Similarly, a pixel $p$ can be back-projected to a 3D point $P$:
\begin{equation}
    \begin{aligned}
    \pi^{-1}(p, d) = \frac{b\cdot f_{x}}{d} \left( \frac{u-c_{x}}{f_{x}}, \frac{v-c_{y}}{f_{y}}, 1 \right)^{\top},
    \end{aligned}
\end{equation}
where $b$ is the baseline between the left and the right cameras and $d$ the disparity. With the relative extrinsic $\textbf{T}_{j\rightarrow t}\in\mathbb{SE}(3)$ from keyframe $j$ to the current frame $t$, we can update every disparity value $d_j$ of keyframe $j$ as: 
\begin{equation}
    \begin{aligned}
    d_{j}^{proj} = \frac{b \cdot f_{x}}{Z_{j}}, \;\textrm{with } P_{j}(X_j, Y_j, Z_j) = T_{j\rightarrow t}^{-1} \cdot \pi^{-1}(p, d_{j}), 
    \end{aligned}
    \label{eq:back-project}
\end{equation}
where $d_{j}^{proj}$ is the updated disparity value. At this point, we obtain the coordinates of $d_{j}^{proj}$ in $t$ through forward warping. To preserve end-to-end requirement, we use the differentiable Softmax Splatting $\overrightarrow{\sigma}$~\cite{softsplat}:
\begin{equation}
    \begin{aligned}
    \hat{d}_{j}^{proj} &=
    \overrightarrow{\sigma}(d_{j}^{proj}, \pi(P_{j}) - p )
    \end{aligned}
    \label{eq:forward-warp}
\end{equation}
Finally, the Local Map is defined as follows:
\begin{equation}
    \begin{aligned}
    \mathcal{D}^{2} = \oplus \left\{ \mathcal{D}^{2}, \hat{d}_{1}^{proj}, \hat{d}_{2}^{proj}, \ldots, \hat{d}_{N_{key}}^{proj} \right\}
    \end{aligned}
\end{equation}

It is worth noticing that, in single-pair mode, the Local Map only contains candidates from the current pair. Moreover, it boosts candidate selection only for $s=2$, since this stage represents the best trade-off between accuracy and speed. In fact, in $s=1$ candidates are looked over the full search range at the lowest resolution, while in $s=3$ a larger cost volume involves a much more expensive computation. 

\textbf{Temporal shift and past costs.}
Past semantic and matching scores are crucial for temporal stereo processing. Although Local Map effectively proposes previous depth cues about the scene, it lacks in providing the context behind these guesses. To address the problem, we introduce Temporal Shift and Past Costs modules. Temporal Shift enriches current feature maps with those computed in the past. Specifically, we adopt the TSM \cite{tsm} strategy to facilitate the feature exchange among neighboring frames because it does not introduce additional computation or parameters: past features are \textit{shifted} along the time dimension and then merged with current ones as shown in~\cref{fig:framework}. Doing so, TSM provides \textit{spatio-temporal} capabilities to our backbone not originally designed for temporal modeling. In practice, every feature map from the backbone is cached and used by Temporal Shift in the next prediction. Notably, as TSM does not rely on pose, we can still perform stereo matching in temporal mode and benefit from past information when pose is not available as shown in~\cref{tab:ablation-temporal}. 

Similarly, Past Costs module adds past matching scores to current cost volumes. Given  $\mathcal{C}_f^3$ volume computed at time $t-1$, first we update the value of its candidates to $t$ according to Eq. (\ref{eq:back-project}). Then, cost values and candidates are forward warped to $t$ using Eq. (\ref{eq:forward-warp}). Finally, the warped cost volume is downsampled by a factor of 2 and 4 and concatenated with current $\mathcal{C}_{agg}^{2}$ and $ \mathcal{C}_{agg}^{1}$ respectively, to be further aggregated by the \nonlocal module. To preserve computation, we only warp top-$K$ candidates and their costs to $t$. In single-pair mode, both Temporal Shift and Past Costs are the identity function. This strategy allows us to run the model trained in temporal mode also in single-pair mode with a limited drop in accuracy, e.g., at bootstrap.

\section{Experiments}

\begin{table}[t]
    \centering
    \resizebox{1.0\columnwidth}{!}{
    \begin{tabular}{cc}
    
    \multicolumn{2}{c}{
        \begin{tabular}{cccc|ccc|ccc|c}
        \hline
        \toprule
            & Multi-Level & Statistical & Adaptive & \multicolumn{3}{c}{EPE} & \multicolumn{3}{c}{3PE} & Runtime \\
            \cline{5-11}
            & Cost & Fusion & Shifting & ALL & OCC & NOC & ALL & OCC & NOC & (ms)  \\
            \toprule
            (A) & \xmark & \xmark & \xmark & 0.600 & 1.949 & 0.369 & 2.85 & 11.62 & 1.36 & \textbf{40.62} \\
            (B) & \cmark  & \xmark & \xmark & 0.587 & 1.924 & 0.360 & 2.79 & 11.44 & 1.33 & 43.41  \\
            (C) & \cmark  & \cmark & \xmark & 0.581 & 1.908 & 0.356 & 2.78 & 11.39 & \textbf{1.31} & 44.54 \\
            (D) & \xmark & \xmark & \cmark & 0.564 & 1.923 & 0.334 & 2.87 & 11.78 & 1.36 & 41.58 \\
            (E) & \cmark  & \cmark & \cmark& \textbf{0.532} & \textbf{1.830} & \textbf{0.315} & \textbf{2.75} & \textbf{11.37} & \textbf{1.31} & 45.42 \\
            \bottomrule
            
        \end{tabular} } \\ 
    \\
    \resizebox{0.90\columnwidth}{!}{
        \begin{tabular}{ccc|c|c}
        \hline
        \toprule
        & Candidates & Disparity & \multirow{2}{*}{EPE} & Runtime \\
            \cline{5-5}
            & Number & Range & & (ms)  \\
            \toprule
            (F) & 3 & [-4, 4] & 0.565 & \textbf{41.61} \\
            (G) & 9 & [-4, 4] & \textbf{0.531} & 55.84  \\
            (H) & 5 & [-4, 4] & 0.532 & 45.42  \\
            (I) & 5 & [-2, 2] & 0.543  & 45.42 \\
            (J) & 5 & [-8, 8] & 0.568 & 45.42 \\
            \bottomrule
        \end{tabular}        
    } &
    \resizebox{1.03\columnwidth}{!}{
        \begin{tabular}{ccc|c|c}
        \hline
        \toprule
        & Model & Depth-wise & \multirow{2}{*}{EPE} & Runtime \\
            \cline{5-5}
            & Variants & 3DCNN & & (ms)  \\
            \toprule
            (K) & Baseline & \xmark & 0.589 & 63.91 \\
            (L) & Baseline & \cmark & 0.600 & \textbf{40.62}  \\
            (M) & Full & \xmark & 0.535 & 75.42  \\
            (N) & Full & \cmark & \textbf{0.532} & 45.42 \\
            \bottomrule
        \end{tabular}        
    }
    \\
    
    \end{tabular}
    }
    \caption{\textbf{Single-pair mode ablation.} We assess on SceneFlow the key components of the proposed architecture.}
    \label{tab:ablation-single}
\end{table}

This section describes the experimental setups used to evaluate on popular datasets, including: SceneFlow~\cite{sceneflow}, TartanAir~\cite{TartanAir}, KITTI 2012~\cite{KITTI2012} and KITTI 2015~\cite{KITTI2015}.
As standard practice in this field~\cite{AcfNet,GANet}, we compute the end-point-error (EPE) and the percentage of points with a disparity error $>3$ pixels (3PE, $>5$ for 5PE) as error metrics in non-occluded (NOC), occluded (OCC) and both (ALL) the regions. Moreover, following the literature~\cite{KITTI2015}, we measure the D1 error on KITTI instead of the 3PE in background (BG), foreground (FG), and both (ALL) areas. We do so by computing the percentage of points with error $>3$ pixels and $>5\%$ than the ground truth. Runtime reported in Tab~\ref{tab:ablation-single},~\ref{tab:ablation-temporal},~\ref{tab:benchmark} is measured in the corresponding image resolution on each dataset, on a single NVIDIA 3090 GPU. The \supp{supplementary material} outlines more details about implementation, dataset description, and training setups.

\begin{table}[t]
    \centering
    \resizebox{0.9\columnwidth}{!}{
        \begin{tabular}{cccc|ccc|ccc|c}
        \hline
        \toprule
             & Temporal & Local & Past & \multicolumn{3}{c}{EPE} & \multicolumn{3}{c|}{3PE} & Runtime \\
            \cline{5-7}  \cline{8-10}
            & Shift & Map & Costs & ALL & OCC & NOC & ALL & OCC & NOC & (ms)  \\
            \toprule
            (A) & \xmark & \xmark & \xmark & 0.647 & 1.899 & 0.420 & 3.96 & 14.21 & 2.08 & \textbf{36.85} \\
            (B) & \cmark & \xmark & \xmark & 0.643 & 1.842 & 0.435 & 4.00 & 14.55 & 2.15 & \textbf{36.85}\\
            (C) & \cmark & \cmark & \xmark & 0.624 & 1.799 & \textbf{0.413} & 3.81 & 13.60 & \textbf{2.02} & 38.22 \\
            (D) & \cmark & \cmark & \cmark & \textbf{0.610} & \textbf{1.637} & \textbf{0.413} & \textbf{3.73} & \textbf{12.60} & 2.03 & 40.13 \\
            \bottomrule
           
        \end{tabular}         

    }
    \caption{\textbf{Temporal mode ablation.} We evaluate the temporal components on TartanAir, with $W_{tr}=W_{test}=4$.}
    \label{tab:ablation-temporal}
\end{table}

\subsection{Ablation Study}

\textbf{Single-pair mode.} We leverage SceneFlow~\cite{sceneflow} to assess the impact of main components of \net{} in single-pair mode. \cref{tab:ablation-single} reports this ablation study, witnessing how each module helps to improve the results consistently. In particular, the multi-level cost computation with group-wise convolutions (B) is effective in enriching the cost volume of each stage, and the baseline model (A) also benefits from the further aggregation of \nonlocal (C). Nonetheless, the adaptive shifting strategy (E) provides without any doubt the main boost in performance. To be noticed, without the enriched context of each sparse candidate, the adaptive shifting strategy shows limitations on disparity correcting and its performance gain (D) decreases a lot. Furthermore, results shown in (F, G, H) illustrates our \net{} is able to predict accurate disparity with as few as 5 candidates. With the ability to shift the candidates towards a better solution, our model can search in a quite large space (e.g., $\beta=2,\; 4, \; 8$ as reported in H, I, J) to get the best result when $\beta=4$. 3D convolutions are the common operations to aggregate the cost in recent methods~\cite{coex,StereoNet,AcfNet}. Our baseline model (A) benefits from the 3D convolutions (K) when compared to depth-wise~\cite{mobilenetv2} 3D convolutions (L), but the runtime increases a lot. In contrast, the full model (E) gives much higher improvement (N) with negligible runtime increase (4.8ms), and the time-consuming 3D convolutions are not necessary (M) for accuracy improvement.

\noindent\textbf{Temporal mode.}
Before presenting the results achieved in temporal mode, we illustrate the protocol adopted at training and test time on TartanAir dataset~\cite{TartanAir}. \textit{Given a temporal window containing $W$ frames}, initial $f=\{1,2,...,W-1\}$ frames are processed by the network sequentially without computing error metrics and blocking the gradients. Specifically, each outcome is cached and used in the next frame prediction. When $f=W$, we compute errors (and backpropagation at training time). Tab.~\ref{tab:ablation-temporal} reports the ablation conducted on TartanAir~\cite{TartanAir}, with $W=4$ both at train and test time, aimed at evaluating the importance of each temporal module. Specifically, we train the model in the single-pair mode for $20$ epochs (i.e., the model learns how to solve stereo matching task), then we enable temporal components for 20 more epochs (i.e., the model now focuses on how to use past information). We can notice how the baseline (A), i.e., the model trained in single-pair mode for 40 epochs, could be improved using Temporal Shift to fuse past features with current ones (B). However, the benefit due to Temporal Shift is much lower than the gain provided by Local Map, which largely improves the performance (C). Finally, including cached past costs as well (D) provides an additional boost in accuracy.

\begin{figure}[t]
    \centering
    \resizebox{0.92\columnwidth}{!}{

    \renewcommand{\tabcolsep}{1pt}   
    \begin{tabular}{cccc}
        Inputs & (1) single-pair & (2) temporal & (3) temporal w/ bad poses \\
        \includegraphics[width=0.22\textwidth]{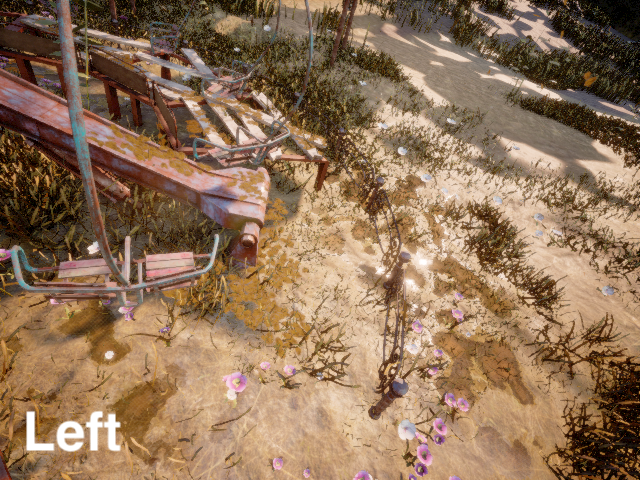} &
        \includegraphics[width=0.22\textwidth]{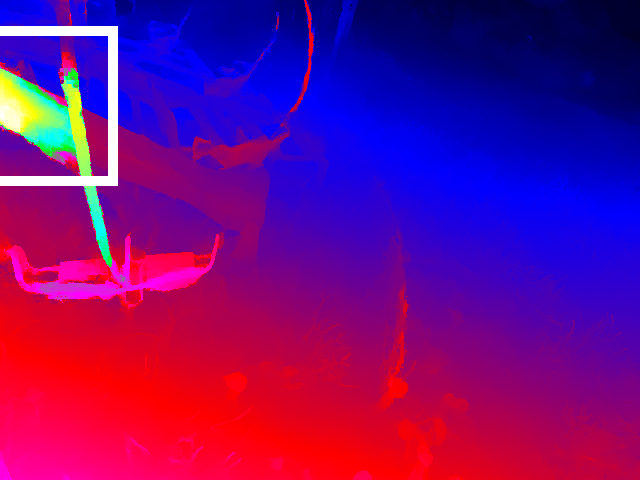} & 
        \includegraphics[width=0.22\textwidth]{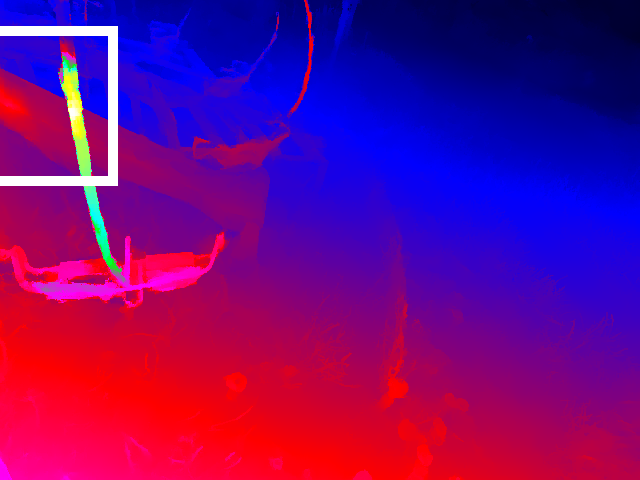} &
        \includegraphics[width=0.22\textwidth]{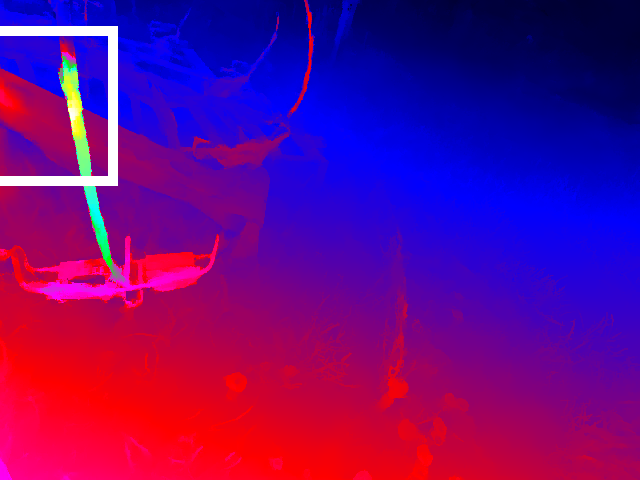}
        \\
        \includegraphics[width=0.22\textwidth]{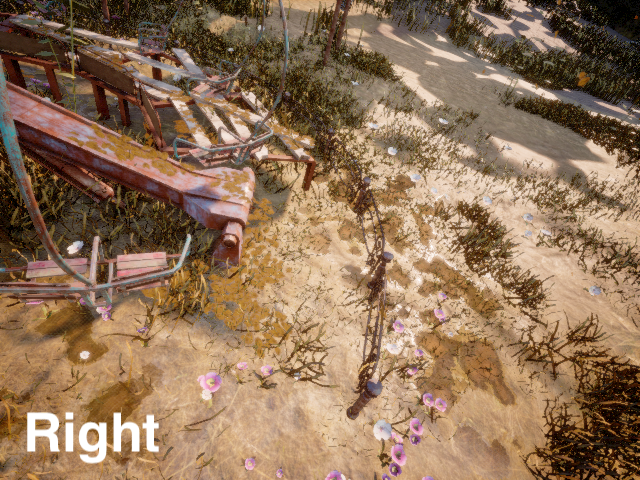} &
        \includegraphics[width=0.22\textwidth]{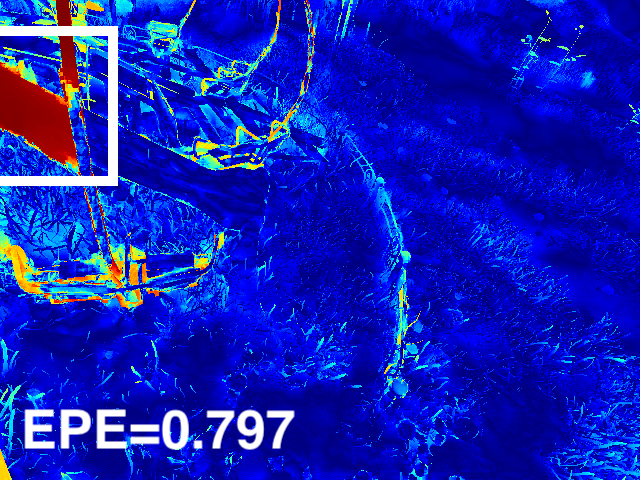} &
        \includegraphics[width=0.22\textwidth]{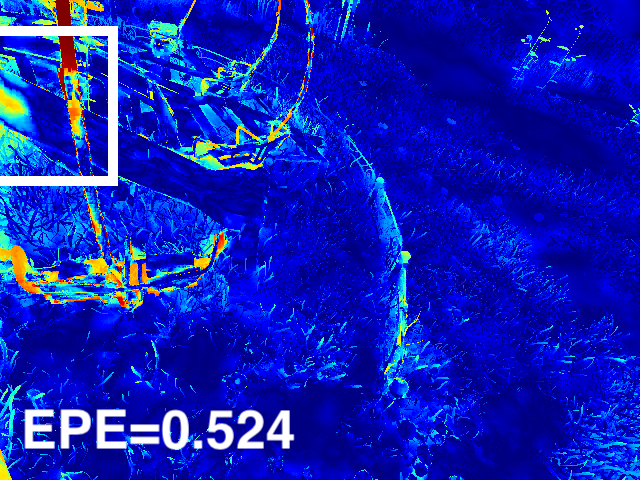} &
        \includegraphics[width=0.22\textwidth]{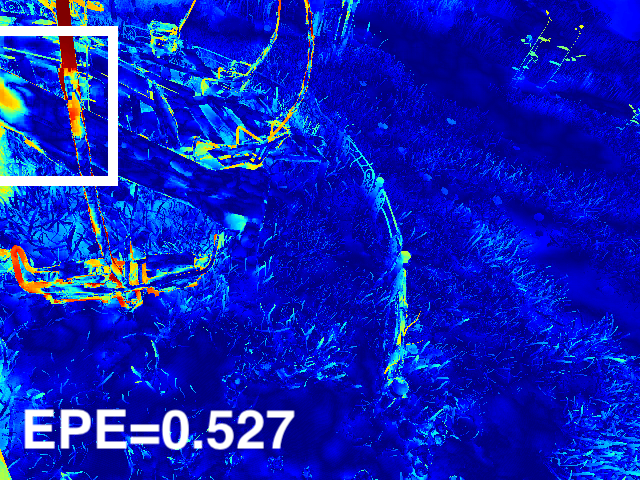} \\
    \end{tabular}
    }
    \caption{\textbf{Benefits of temporal mode.} Compared to single-pair mode (1), temporal mode (2) is more accurate at occlusions, even with noisy poses (3) -- colder colors encode lower error.
    }
    \label{fig:qualitative-tartan}
\end{figure}

\textbf{Impact of temporal window.} Tab. \ref{tab:ablation-windows} reports the results achieved by \net{} using different values of $W$. In particular, we could have two different values for $W$, that are $W_{tr}$ and $W_{test}$ for train and test respectively. In addition to the baseline $W_{tr}=1$ and the temporal $W_{tr}=4$ models, we also train a $W_{tr}=8$ model following the same configuration as for $W_{tr}=4$. When compared with existing models such as PSMNet~\cite{PSMNet} and CoEx~\cite{coex}, our baseline outperforms them by a large margin in EPE metric. As for temporal mode, in general, the more frames joining the training or testing phases, the better result we can get in all regions. 
Notably, \colorbox{w4!60!white}{$W_{test}=4$} and \colorbox{w8!60!white}{$W_{test}=8$} are always beneficial in OCC, witnessing that \net{} can effectively exploit more frames to deal with such difficult areas. 
Moreover, when $W_{test} > 1$, the results consistently overcome baseline with single stereo pair. It indicates our network can benefit from past information with only a few frames (e.g., 4, 8) after the network startup. Finally, we can notice how temporal models tested with \colorbox{w1!60!white}{$W_{test}=1$} obtain results closer to the baseline. This outcome implies that, in practical video applications, \net{} can be trained once in temporal mode and used in single-pair way for the first inference (yet providing a good initial estimate) and in the temporal mode for all the others. \cref{fig:qualitative-tartan} visualises the benefits of temporal model ($W_{tr}=8$, \colorbox{w8!60!white}{$W_{test}=8$}) to alleviate occlusion errors. The temporal mode (2) largely outperforms the single-pair one (1). Furthermore, we highlight how \net{} is also robust against inaccurate poses: although in (3) the camera pose is set to an identity matrix and only Temporal Shift module could help, it still outperforms (1). Eventually, the proposed temporal cues can be plugged into recent efficient methods (e.g., CoEx~\cite{coex} and StereoNet~\cite{StereoNet}) for further improvement. Although CoEX with a very different architecture compared to ours, its EPE still decreases from 0.714 to 0.610, with near 14\% improvement. Nonetheless, our \net{} utilizes past cues more effectively.

\begin{table}[t]
    \centering
    \resizebox{0.9\columnwidth}{!}{
    \begin{tabular}{clc|ccc|ccc}
        \toprule
        & \multirow{2}{*}{Method} & \multirow{2}{*}{$W_{test}$} & \multicolumn{3}{c}{EPE} & \multicolumn{3}{c}{3PE} \\
        \cline{4-9} & & & ALL & OCC & NOC & ALL & OCC & NOC \\
        
        \midrule
        
        \multirow{4}{*}{\rotatebox[origin=c]{90}{$W_{tr}=1$}} & PSMNet~\cite{PSMNet} & \cellcolor{w1!60!white}1 & \cellcolor{w1!60!white}0.866 & \cellcolor{w1!60!white}2.654 & \cellcolor{w1!60!white}0.558 & \cellcolor{w1!60!white}4.80 & \cellcolor{w1!60!white}18.63 & \cellcolor{w1!60!white}2.48 \\
        & StereoNet~\cite{StereoNet} & \cellcolor{w1!60!white}1 & \cellcolor{w1!60!white}0.888 & \cellcolor{w1!60!white}2.647 & \cellcolor{w1!60!white}0.578 & \cellcolor{w1!60!white}5.15 & \cellcolor{w1!60!white}19.34 & \cellcolor{w1!60!white}2.68 \\
        & CoEx~\cite{coex} & \cellcolor{w1!60!white}1 & \cellcolor{w1!60!white}0.714 & \cellcolor{w1!60!white}2.074 & \cellcolor{w1!60!white}0.463 & \cellcolor{w1!60!white}\textbf{3.83} & \cellcolor{w1!60!white}14.57 & \cellcolor{w1!60!white}\textbf{1.93} \\
        & Ours (single-pair) & \cellcolor{w1!60!white}1 & \cellcolor{w1!60!white}\textbf{0.647}  & \cellcolor{w1!60!white}\textbf{1.899} & \cellcolor{w1!60!white}\textbf{0.420}  & \cellcolor{w1!60!white}3.96  &  \cellcolor{w1!60!white}\textbf{14.21} & \cellcolor{w1!60!white}2.08 \\
        
        \midrule
        \multirow{3}{*}{\rotatebox[origin=c]{90}{$W_{tr}=4$}} & Ours (single-pair) & \cellcolor{w1!60!white}1 & \cellcolor{w1!60!white} 0.665 & \cellcolor{w1!60!white} 1.731 & \cellcolor{w1!60!white} 0.459 & \cellcolor{w1!60!white} 3.95 & \cellcolor{w1!60!white} 13.34 & \cellcolor{w1!60!white} 2.16 \\
        
        & Ours (temporal) & \cellcolor{w4!60!white}4 & \cellcolor{w4!60!white}0.610 & \cellcolor{w4!60!white}1.637& \cellcolor{w4!60!white}0.413 & \cellcolor{w4!60!white}3.73 & \cellcolor{w4!60!white}12.60 & \cellcolor{w4!60!white}2.03 \\
        
        & Ours (temporal) & \cellcolor{w8!60!white}8 & \cellcolor{w8!60!white}\textbf{0.607} & \cellcolor{w8!60!white}\textbf{1.634} & \cellcolor{w8!60!white}\textbf{0.409} & \cellcolor{w8!60!white}\textbf{3.71} & \cellcolor{w8!60!white}\textbf{12.59} & \cellcolor{w8!60!white}\textbf{2.01} \\
        
        \midrule
        
        \multirow{5}{*}{\rotatebox[origin=c]{90}{$W_{tr}=8$}} & Ours (single-pair) & \cellcolor{w1!60!white}1 & \cellcolor{w1!60!white} 0.673 & \cellcolor{w1!60!white} 1.912 & \cellcolor{w1!60!white} 0.450 & \cellcolor{w1!60!white} 4.00 & \cellcolor{w1!60!white} 14.03 & \cellcolor{w1!60!white} 2.17 \\
        
        & Ours (temporal) & \cellcolor{w4!60!white}4 & \cellcolor{w4!60!white}0.609  & \cellcolor{w4!60!white}1.625 & \cellcolor{w4!60!white}0.412 & \cellcolor{w4!60!white}3.74 & \cellcolor{w4!60!white}12.58 & \cellcolor{w4!60!white}2.03\\
        
        & StereoNet~\cite{StereoNet} (temporal) & \cellcolor{w8!60!white}8 & \cellcolor{w8!60!white}0.656 & \cellcolor{w8!60!white}1.928 & \cellcolor{w8!60!white}0.428 & \cellcolor{w8!60!white}3.97 & \cellcolor{w8!60!white}14.45 & \cellcolor{w8!60!white}2.06 \\
        
        & CoEx~\cite{coex} (temporal) & \cellcolor{w8!60!white}8 & \cellcolor{w8!60!white}0.610 & \cellcolor{w8!60!white}1.637 & \cellcolor{w8!60!white}0.413 & \cellcolor{w8!60!white}3.73 & \cellcolor{w8!60!white}12.60 & \cellcolor{w8!60!white}2.03 \\
        
        & Ours (temporal) & \cellcolor{w8!60!white}8 & \cellcolor{w8!60!white}\textbf{0.601} & \cellcolor{w8!60!white}\textbf{1.615} & \cellcolor{w8!60!white}\textbf{0.405} & \cellcolor{w8!60!white}\textbf{3.71} & \cellcolor{w8!60!white}\textbf{12.54} & \cellcolor{w8!60!white}\textbf{2.02} \\
        \bottomrule
        \end{tabular}
    }
    \caption{\textbf{Impact of different frames in temporal mode.} Models are tested with $W_{test}=$ 1, 4 and 8.
    }

	\label{tab:ablation-windows}
\end{table}

\begin{table}[t]
    \centering
    \resizebox{0.9\columnwidth}{!}{
    \begin{tabular}{ccc|cc|cc|cc|cc}
        \toprule
        \multicolumn{3}{c|}{\multirow{3}{*}{Pose Type}} & \multicolumn{4}{c}{\cellcolor{lightwhite!60!white} Outdoor} & \multicolumn{4}{c}{\cellcolor{lightblue!60!white}Indoor}\\
        &&& \multicolumn{2}{c}{\cellcolor{lightwhite!60!white}Amusement} 
        & \multicolumn{2}{c}{\cellcolor{lightwhite!60!white}SoulCity}
        &\multicolumn{2}{c}{\cellcolor{lightblue!60!white}Carwelding} &
        \multicolumn{2}{c}{\cellcolor{lightblue!60!white}Hospital} \\
        \cline{4-11}
        &&& EPE & 3PE & EPE & 3PE & EPE & 3PE & EPE & 3PE \\
        \midrule
        \multicolumn{3}{c|}{Single-pair} & 0.562 & 2.62 & 0.529 & 2.60 & 0.508 & 2.84 & 0.904 & 3.74   \\
        \multicolumn{3}{c|}{Identity} & 0.569 & 2.58 & 0.534 & 2.66 & 0.538 & 2.96 & 0.927 & 4.21 \\
        \multicolumn{3}{c|}{DROID-SLAM~\cite{teed2021droid}} & \textbf{0.545} & \textbf{2.51} & 0.509 & \textbf{2.52} & 0.479 & \textbf{2.63} & \textbf{0.848} & \textbf{3.30} \\
        \multicolumn{3}{c|}{GT} & \textbf{0.545} & \textbf{2.51} & \textbf{0.508} & \textbf{2.52} & \textbf{0.478}  & \textbf{2.63} & \textbf{0.848} & \textbf{3.30} \\
        \cline{1-11}
        & $\sigma_{R}$ $(^{\circ})$ & $\sigma_{t}$ $(m)$ \\
        \cline{1-11}
        \multirow{5}{*}{\rotatebox[origin=c]{90}{GT+Noise}} & 10 & 0.50 & 0.716 & 3.52 & 0.791 & 4.86 & 0.608 & 3.34 & 1.152 & 4.85 \\
        & 10 & 0.05 & 0.713 & 3.46 & 0.776 & 4.55 & 0.611 & 3.37 & 1.096 & 4.79 \\
        & 1 & 0.50 & 0.577 & 2.58 & 0.592 & 2.83 & 0.531 & 2.91 & 0.934 & 4.49  \\
        & 1 & 0.05 & \textbf{0.546} & \textbf{2.54} & 0.518 & 2.64 & 0.506 & \textbf{2.72} & 0.880 & 3.84 \\
        & 1 & 0.01 & \textbf{0.546} & \textbf{2.54} & \textbf{0.517} & \textbf{2.62} & \textbf{0.505} & 2.74 & \textbf{0.873} & \textbf{3.66}  \\
        \bottomrule
        \end{tabular}
    
    }
    \caption{\textbf{Impact of different pose input in temporal mode.} ``Single-pair" denotes our model is in single-pair mode (no pose needed); ``Identity" means identity matrix; ``DROID-SLAM" means pose estimated by DROID-SLAM (ORB-SLAM3 fails on these 4 scenes);
    and ``GT+Noise" is the ground truth pose with manually added Gaussian noise.
    }
	\label{tab:ablation-pose}
\end{table}

\textbf{Pose Analysis.} To assess the impact of pose input, we evaluate our model with setting $W_{tr}=8$ and \colorbox{w8!60!white}{$W_{test}=8$} on 4 scenes (2 indoor and 2 outdoor scenes) of TartanAir dataset~\cite{TartanAir} with hard motion patterns. 
The overall results of several pose inputs
are listed in Tab~\ref{tab:ablation-pose}. Specifically, the noise of rotation follows $\mathcal{N}(0, \sigma_{R})$ in degree($^\circ$) and noise of translation follows $\mathcal{N}(0,\sigma_{t})$ in meter(m). As reported, 1) poses from ground truth (GT) or estimated by DROID-SLAM~\cite{teed2021droid} yield almost the same EPE and 3PE metrics. It means, when the ground truth pose is not available, an actual SLAM system like DROID-SLAM could be an alternative scheme. 2) Pose with small rotation error (e.g., $\sigma_{R} <= 1^{\circ}$) and translation error (e.g., $\sigma_{t} <= 0.05m$) gets very close results to the one with ground truth pose, and it always surpasses the results by single-pair mode. Even when the pose error comes to the maximum level of the dataset (i.e., $\sigma_{R} = 10^{\circ}, \sigma_{t} = 0.5m$), our model does not crash down and provides impressive predictions, proving the robustness of our model to inaccurate pose. 3) For outdoor scenes, since the view change between frames is much smaller than indoor scenes, the accuracy drops negligibly even when identity transformation is applied. 

\begin{table}[t]

    \centering
    \resizebox{1.0\columnwidth}{!}{
    \begin{tabular}{c|c|c|c|c|c|c|c|c}
    \hline
    \toprule
        Pretrain & \multicolumn{2}{c|}{SF} & \multicolumn{2}{c|}{SF+Pseudo} & \multicolumn{4}{c}{TartanAir+Pseudo} \\
        \hline
        Method & CoEx & Ours & CoEx & Ours & CoEx & Ours & CoEx$\dagger$ & Ours$\dagger$ \\
        \hline
        D1-BG & 1.79 & 2.17  & 1.73  & 1.89 & 1.74 & 1.89 & 1.71 & \textbf{1.61}  \\
        D1-FG & 3.82 & 2.96 & 3.60 & 2.85 & 3.49 & 3.03 & \textbf{2.78} & \textbf{2.78} \\
        D1-ALL & 2.13 & 2.30 & 2.04 & 2.05 & 2.03 & 2.07 & 1.89 & \textbf{1.81} \\
        \bottomrule
       
    \end{tabular}
    }
    \caption{\textbf{Impact of pretraining.} We study the importance of pretraining by evaluating the D1 metric on KITTI 2015 test dataset. Results by both CoEx and Ours in single-pair and temporal mode ($\dagger$) are reported accordingly. 'SF' and 'TartanAir' denotes SceneFlow and TartanAir datasets respectively. `+Pseudo' means further training on KITTI raw sequences with our generated pseudo label.}
    \label{tab:pretraining}
\end{table}
\begin{table}[t]
	\centering
    \resizebox{0.95\columnwidth}{!}{
    \begin{tabular}{cl|c|cc|cc|ccc}
	\toprule
			&\multirow{3}*{Method} & SceneFlow~\cite{sceneflow} & \multicolumn{4}{c|}{KITTI 2012~\cite{KITTI2012}} &\multicolumn{3}{c}{KITTI 2015~\cite{KITTI2015}} \\
			\cline{3-10}
			& & \multirow{2}*{EPE} & \multicolumn{2}{c|}{Reflective} & \multicolumn{2}{c|}{All} &  \multicolumn{3}{c}{D1} \\
			& & & 3PE & 5PE & 3PE & 5PE & BG & FG & ALL \\
			\hline
	   	   \multirow{10}{*}{\rotatebox[origin=c]{90}{slow}}& PSMNet~\cite{PSMNet} & 1.09 & 10.18 & 5.64 & 1.89 & 1.15 & 1.86 & 4.62 & 2.32  \\
			& GwcNet-gc~\cite{GWCNet} & 0.77 & 9.28 & 5.22 & 1.70 & 1.03 & 1.74 & 3.93 & 2.11  \\
			& GANet-Deep~\cite{GANet} & 0.78 & 7.92 & 4.41 & 1.60 & 1.02 & 1.48 & 3.46 & 1.81 \\
			& AcfNet~\cite{AcfNet} & 0.87 & 8.52 & 5.28 & 1.54 & 1.01 & 1.51 & 3.80 & 1.89  \\
			& LEAStereo~\cite{LEAStereo} & 0.78 & \textbf{6.50} & \textbf{3.18} & 1.45 & 0.88 & 1.40 & \textbf{2.91} & \textbf{1.65} \\
			& ACVNet~\cite{ACVNet} & \textbf{0.46} & 7.03 & 4.14 & \textbf{1.13} & \textbf{0.71} & \textbf{1.37} & 3.07 & \textbf{1.65} \\
			& CFNet~\cite{CFNet} & - & 7.29 & 3.81 & 1.58 & 0.94 & 1.54 & 3.56 & 1.88 \\
            \cline{2-10}
            & DWARF$\ddagger$~\cite{aleotti2020learning} & - & - & - & - & - & 3.20 & 3.94 & 3.33 \\
            & DTF\_SENSE$\ddagger$~\cite{schuster2021dtf} & - & - & - & - & - & 2.08 & 3.13 & 2.25 \\
            
            & SENSE$\ddagger$~\cite{SENSE} & - & - & - & - & - & 2.07 & 3.01 & 2.22 \\
            
            \midrule
		    
			\multirow{7}{*}{\rotatebox[origin=c]{90}{fast}} & StereoNet~\cite{StereoNet} & 1.10 & - & - & 6.02 & - & 4.30 & 7.45 & 4.83 \\
			& DeepPruner-Fast~\cite{DeepPruner} & 0.97 & - & - & - & - & 2.32 & 3.91 & 2.59 \\
			& AANet+~\cite{AANet} & 0.72 & 9.10 & 5.12 & 2.04 & 1.30 & 1.65 & 3.96 & 2.03 \\
			& CoEx~\cite{coex} & 0.69 & 8.63 & 4.49 & 1.93 & 1.13 & 1.79 & 3.82 & 2.13 \\
			& HITNet~\cite{HitNet} & \textbf{0.53} & 7.54 & 4.01 & 1.89 & 1.29 & 1.74 & 3.20 & 1.98 \\
			& Ours (single-pair)  & \textbf{0.53} & 6.99 & 3.52 & 1.94 & 1.08 & 1.89 & 2.85 & 2.05 \\
			& Ours (temporal)& - & \textbf{6.14} & \textbf{3.08} & \textbf{1.61} & \textbf{0.88} & \textbf{1.61} & \textbf{2.78} & \textbf{1.81} \\
			
		\bottomrule
    \end{tabular}
    }
	\caption{\textbf{Comparison with state-of-the-art methods -- slow and fast.} We report results of state-of-the-art methods on SceneFlow and KITTI. \textit{fast} denotes models allowing real-time inference. $\ddagger$ means 3D scene flow-based methods.
    }
	\label{tab:benchmark}
\end{table}

\subsection{Evaluations on KITTI Benchmarks}

To conclude, we run TemporalStereo on KITTI 2012 and KITTI 2015 test data and submit to the online leaderboard. 

\textbf{Pretraining.} As KITTI is very challenging due the lack of a big training set, pretraining on SceneFlow dataset~\cite{sceneflow} is a common training schedule for deep learning-based stereo methods~\cite{PSMNet,LEAStereo}. Recent methods~\cite{CFNet,HSM} also introduce extra data, e.g., HR-VS~\cite{HSM}, to augment KITTI during training. Following the knowledge distillation strategy proposed in AANet+~\cite{AANet}, we augment the KITTI dataset by leveraging the prediction results from pretrained LEAStereo~\cite{LEAStereo} to generate pseudo labels on KITTI raw sequences~\cite{eigen2014depth}. As a result, we get 61 stereo video sequences (containing 42K pairs) with pseudo labels for pretraining, and poses are calculated from the GPS/OXTS data on KITTI. We study the influence of different pretraining strategies on the D1 metric on KITTI 2015 test dataset. As shown in Tab.~\ref{tab:pretraining}, compared with CoEx~\cite{coex}, our model in single-pair mode performs better in D1-FG and worse in D1-ALL, D1-BG metrics when pretrained on SceneFlow dataset only. After further training on pseudo labels, we get the almost same result as CoEx, which demonstrates our model requires more data to achieve better performance. Replacing the SceneFlow dataset with TartanAir, both the result of CoEx and ours in single-pair mode do not improve and achieve almost the same accuracy, i.e., D1-ALL 2.03\% for CoEx and 2.07\% for ours. However, by leveraging temporal information, the temporal mode can boost the accuracy of both \net{} and CoEx further by a large margin, i.e. reducing D1-ALL to 1.81\% and 1.89\% respectively, supporting the major impact of our temporal paradigm over the pseudo labels training. We also point out that, despite the poses of KITTI raw sequences and KITTI 2015 are estimated by GPS/OXTS data and ORBSLAM3~\cite{ORBSLAM3} respectively,  
\net{} demonstrates its robustness to the pose error again.

Tab. \ref{tab:benchmark} collects results achieved by a variety of deep stereo models, both on Scene Flow and the KITTI online benchmarks. For KITTI, we report the error rates achieved by TemporalStereo, both in single-pair and temporal mode ($W_{tr}, W_{test}=11$), finetuned from models pretrained on KITTI raw sequences~\cite{eigen2014depth}. In single-pair mode, our \net{} achieves results already on par with state-of-the-art on all datasets. When switching to the temporal mode, our network surpasses all fast stereo architectures by a large margin. In particular, our result on D1-FG is even better than the one achieved by slow models~\cite{LEAStereo,GANet} running in hundreds of milliseconds. It is worth mentioning that \net{} in temporal mode, benefiting from past semantic and geometric information, achieves better results compared to single-pair mode on D1-FG metric (the D1 error on foreground areas, i.e., moving cars), which proves the robustness of our model to \textbf{dynamic objects} as well. The same advantage is also evident on 3PE and 5PE in \textbf{reflective regions} on KITTI 2012.
Finally, we also compare with \textbf{3D scene flow} based state-of-the-art methods~\cite{aleotti2020learning,schuster2021dtf,SENSE}, which project with dense 3D motion field and output disparity. In contrast, simply using camera pose, our \net{} is the obvious winner in both accuracy and efficiency on KITTI 2015.

\section{Conclusions and Limitations}
We presented \net, a novel network devoted to fast stereo matching. The enhanced coarse-to-fine design and sparse cost volumes allow for fast inference and high performance. Moreover, \net{} can exploit past information to ameliorate predictions, especially in occluded regions. The same model, trained once, can handle either single or multiple stereo pairs effectively. Considering the requirement of camera poses as its main limitation, empowering our system with pose estimation will be our future research direction.

\textbf{Acknowledgment.} We sincerely thank the scholarship supported by China Scholarship Council (CSC).

{\small
\bibliographystyle{ieee_fullname}
\bibliography{egbib}
}

\clearpage

\setcounter{section}{0}
\setcounter{figure}{0}
\setcounter{table}{0}
\renewcommand\thesection{\Alph{section}}
\renewcommand\thetable{\Alph{table}}
\renewcommand\thefigure{\Alph{figure}}

\title{\vspace{-0.5em}-- Supplementary Material --\\TemporalStereo: Efficient Spatial-Temporal Stereo Matching Network}

\author{Youmin Zhang \hspace{0.5cm} Matteo Poggi \hspace{0.5cm} Stefano Mattoccia \\
CVLAB, Department of Computer Science and Engineering (DISI) \\
University of Bologna, Italy\\
\texttt{\{youmin.zhang2, m.poggi, stefano.mattoccia\}@unibo.it}
}

\maketitle

\begin{figure*}[htbp]
    \centering
    \resizebox{2.0\columnwidth}{!}{
    
    \renewcommand{\tabcolsep}{1pt}   
    \begin{tabular}{cccc}
        Left & Ground Truth & Prediction & Error \\
        \includegraphics[width=0.3\textwidth]{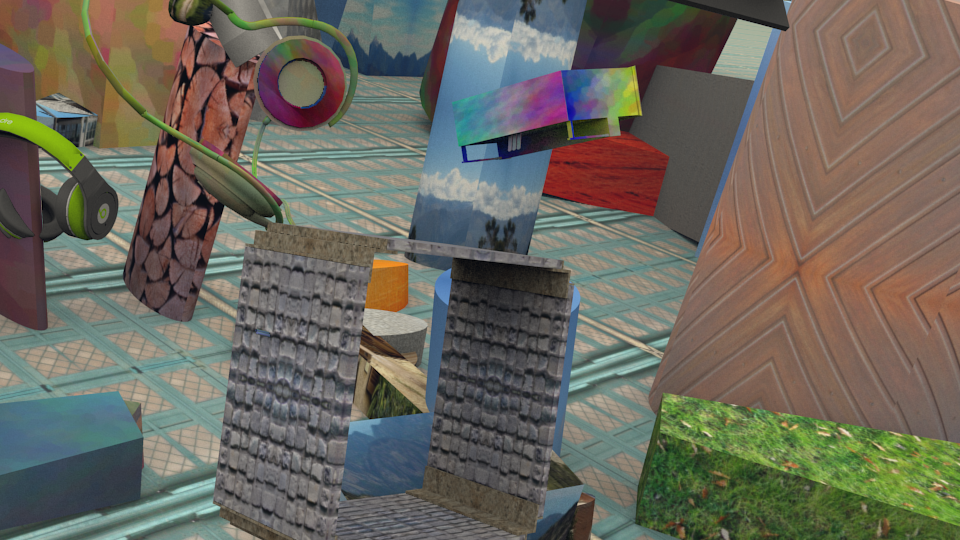}  &
        \includegraphics[width=0.3\textwidth]{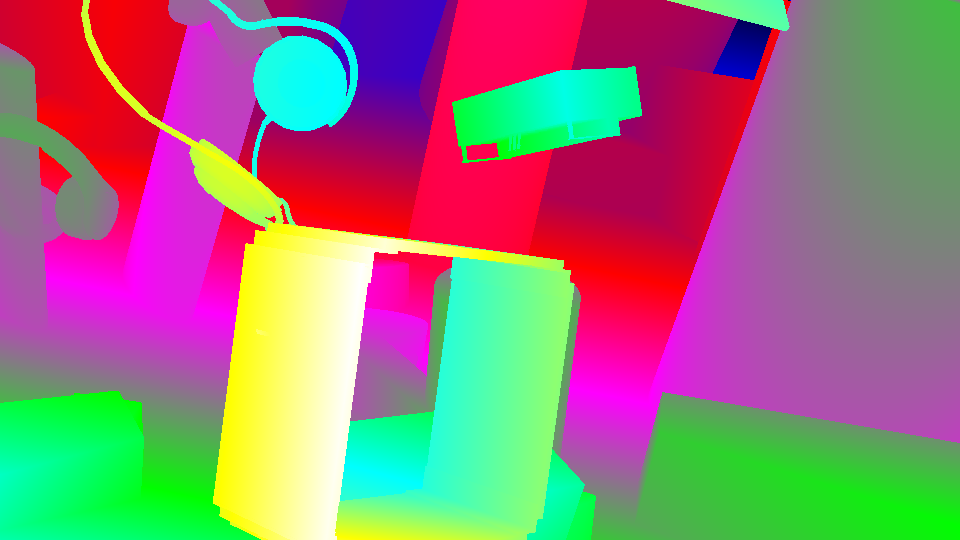} &
        \includegraphics[width=0.3\textwidth]{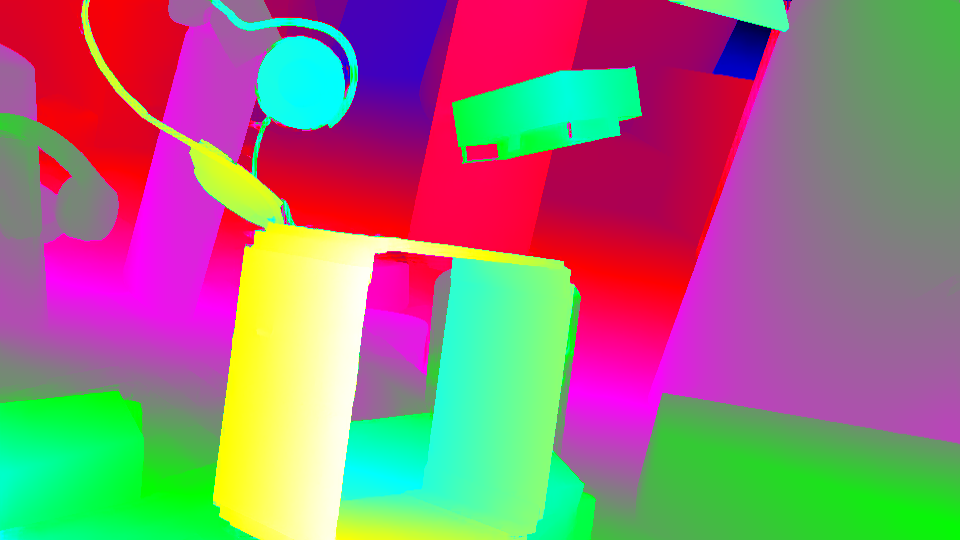} &
        \includegraphics[width=0.3\textwidth]{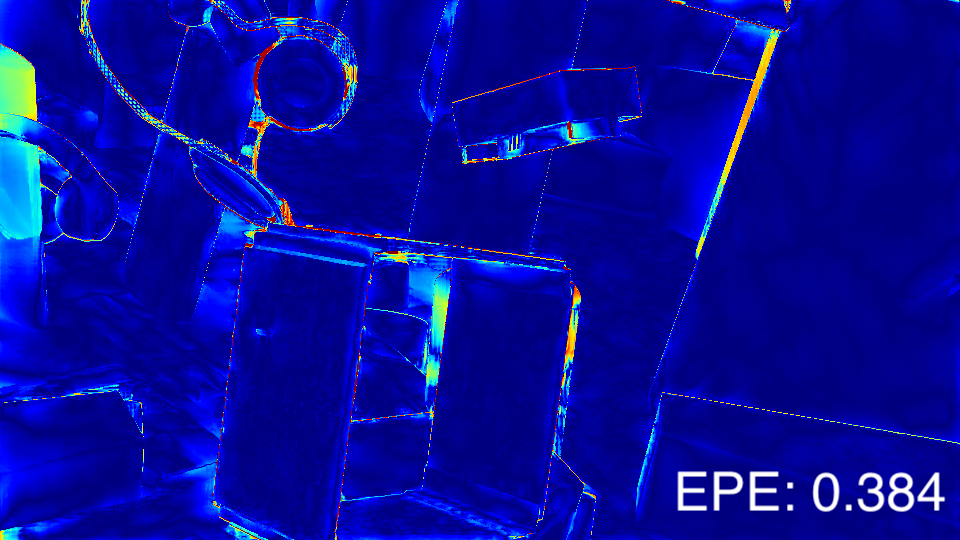}\\
        \includegraphics[width=0.3\textwidth]{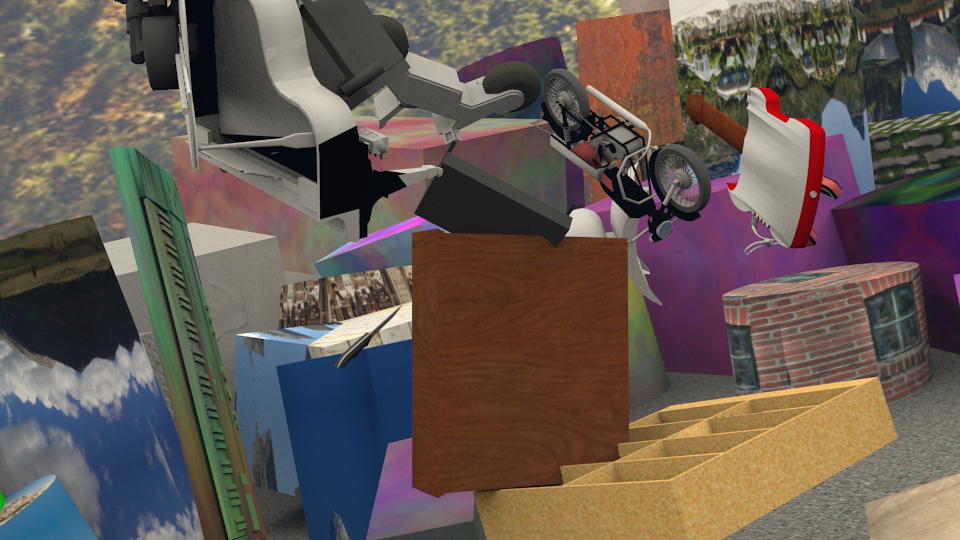}  &
        \includegraphics[width=0.3\textwidth]{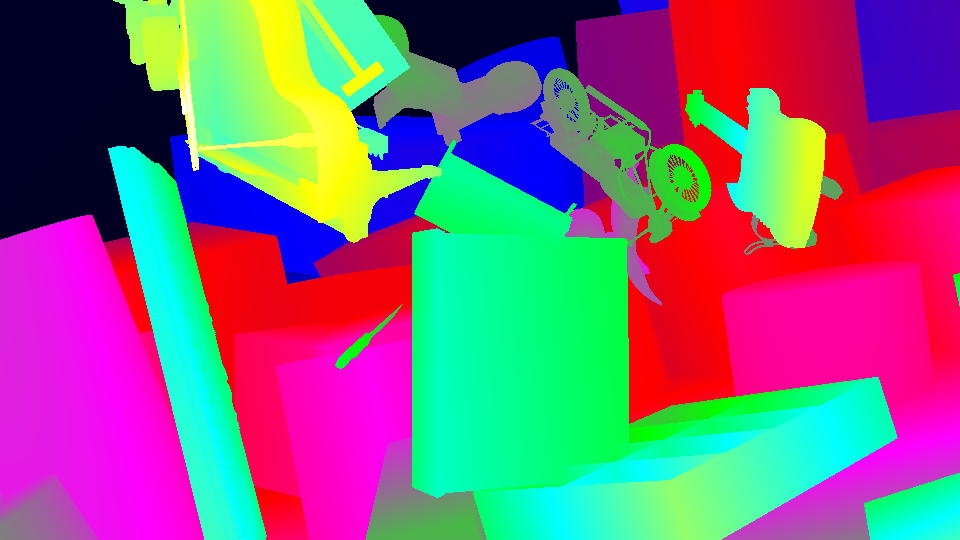} &
        \includegraphics[width=0.3\textwidth]{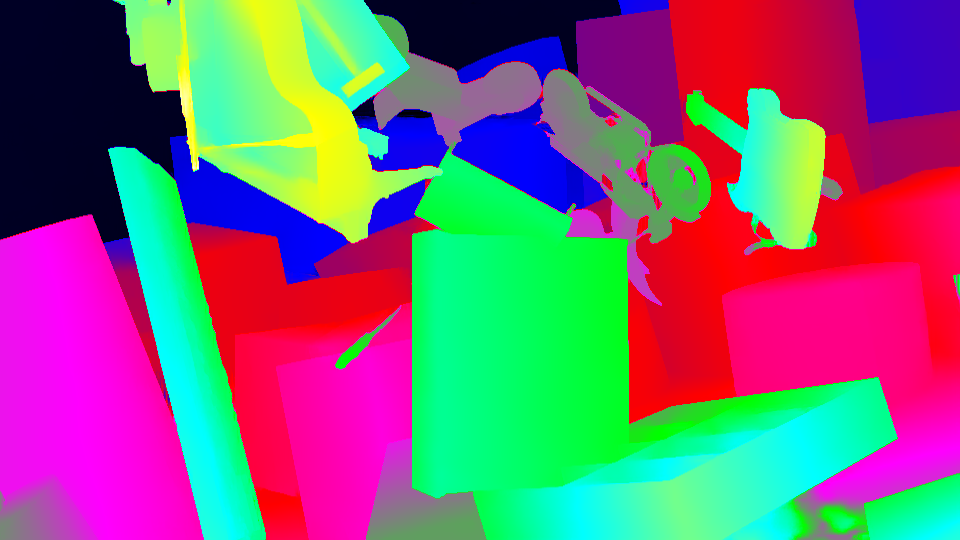} &
        \includegraphics[width=0.3\textwidth]{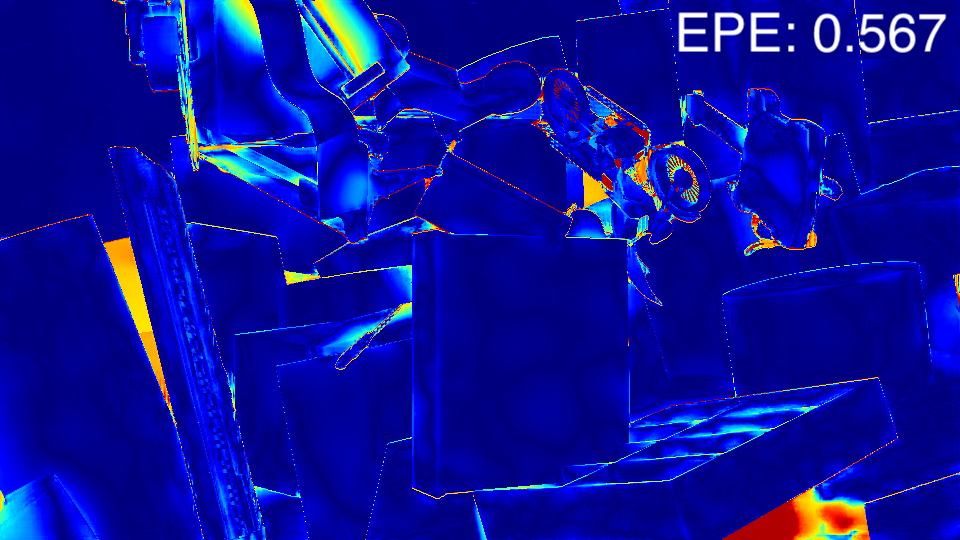}\\
       
    \end{tabular}
    }
    \caption{\textbf{Qualitative results on SceneFlow.} From left to right, the reference/left image, the ground truth disparity map, the prediction in single-pair mode and its error (darker the color, lower the error).}
    \label{fig:qualitative-sceneflow}
\end{figure*}

\begin{figure*}[!h]
    \centering
    \renewcommand{\tabcolsep}{1pt}
    \resizebox{2.0\columnwidth}{!}{

    \begin{tabular}{ccccc}
        Left Image &  CoEx & LEAStereo & Ours(single-pair) & Ours(temporal) \\
        \includegraphics[width=0.24\textwidth]{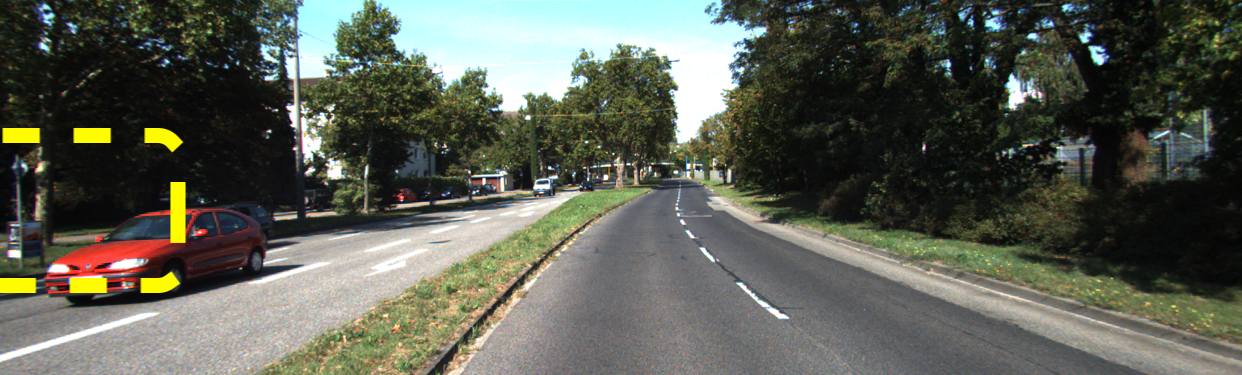} &
        \includegraphics[width=0.24\textwidth]{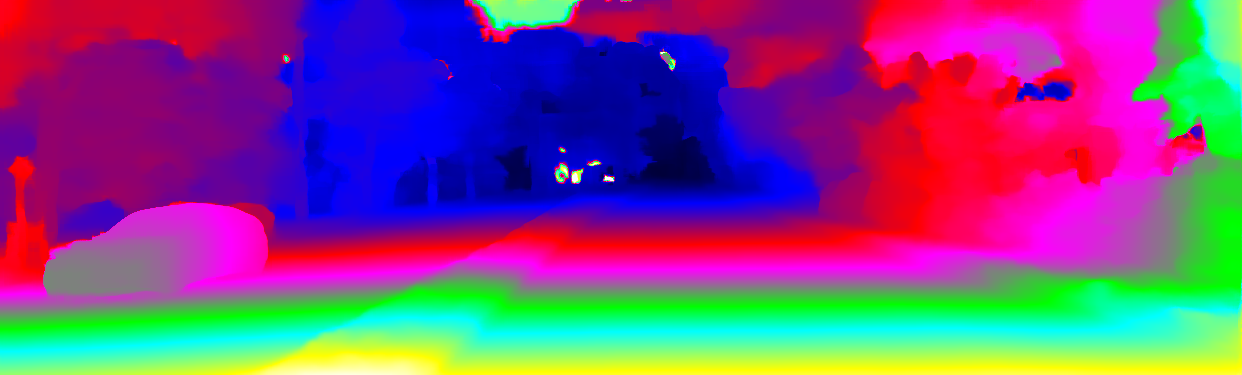} &
        \includegraphics[width=0.24\textwidth]{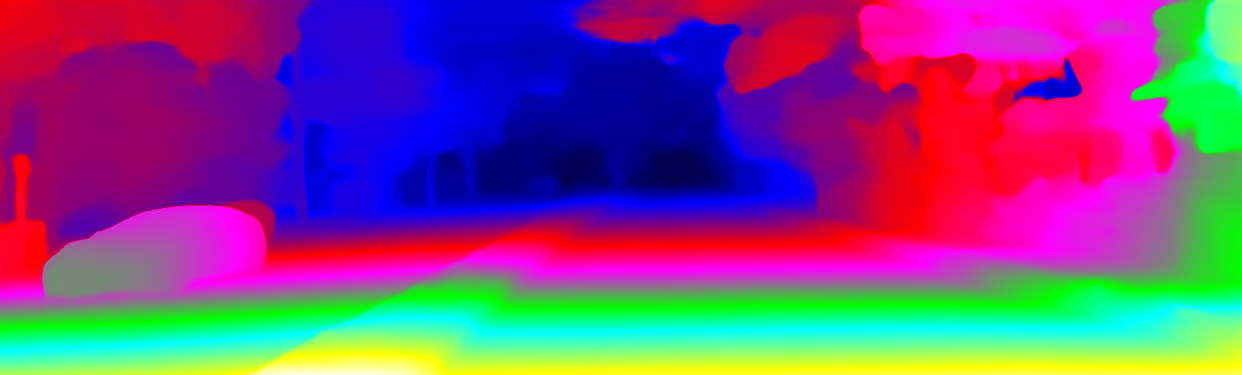} &
        \includegraphics[width=0.24\textwidth]{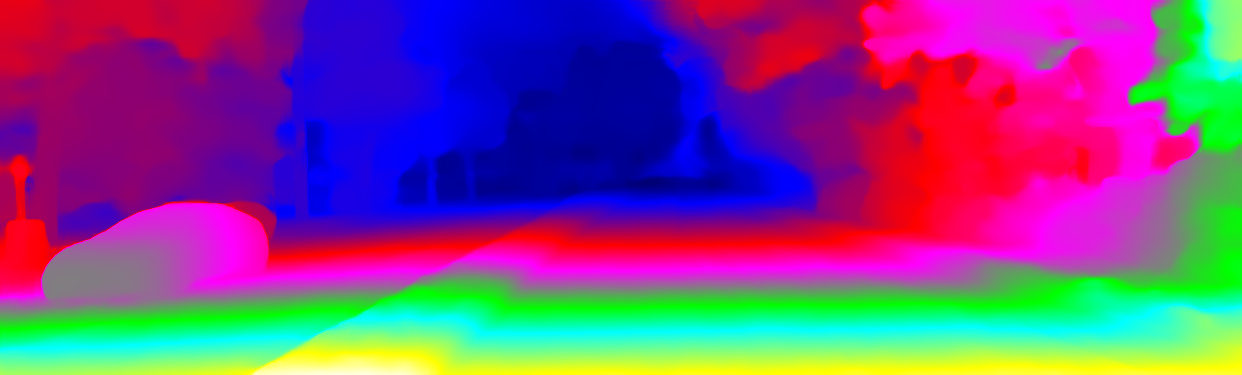} &
        \includegraphics[width=0.24\textwidth]{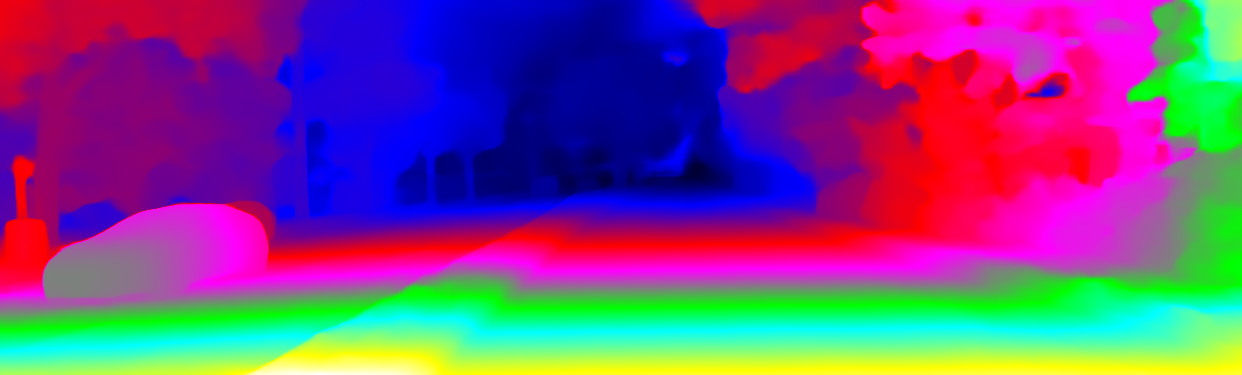} \\
        \includegraphics[width=0.24\textwidth]{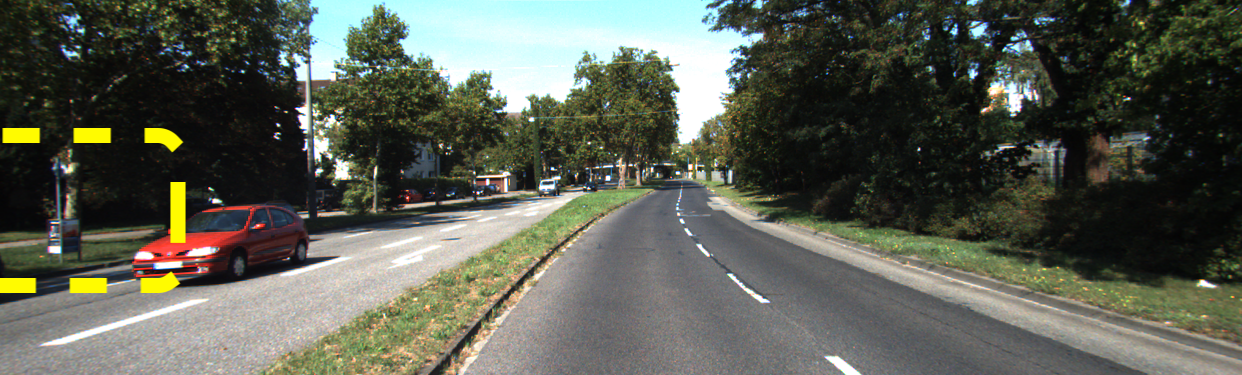} &
        \includegraphics[width=0.24\textwidth]{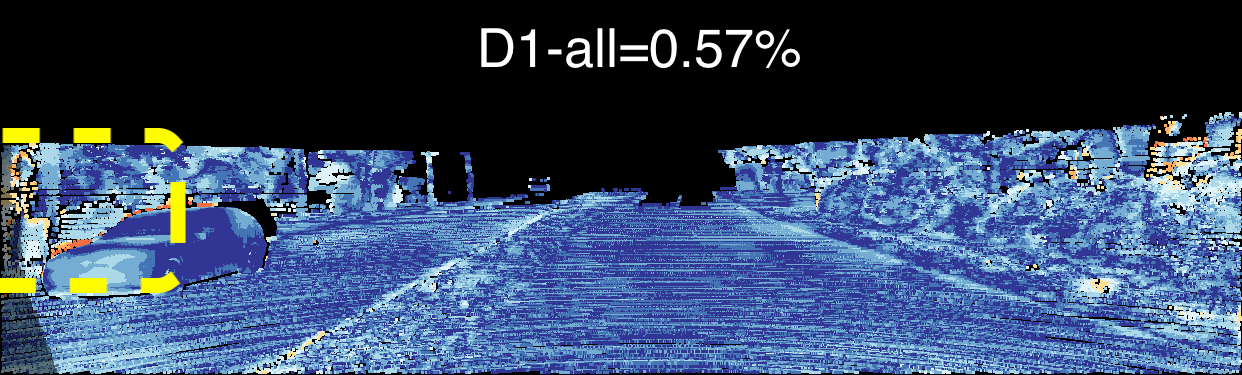} &
        \includegraphics[width=0.24\textwidth]{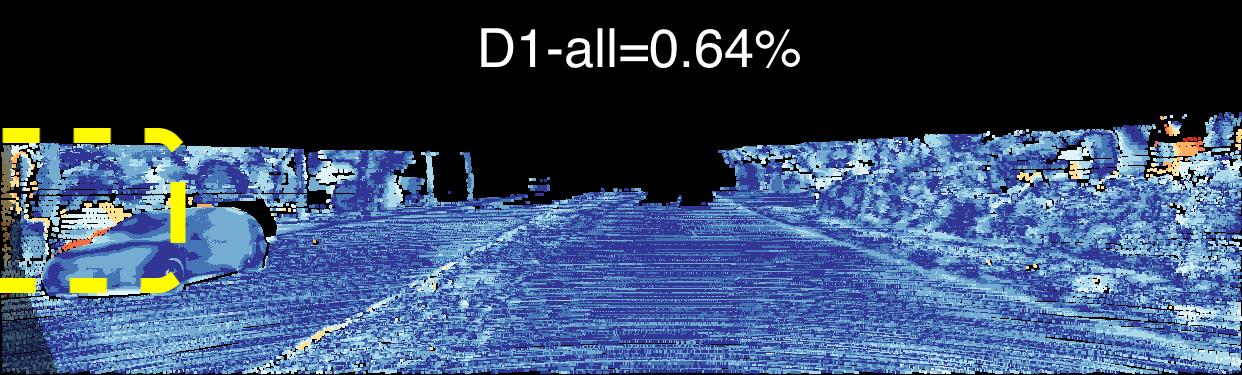} &
        \includegraphics[width=0.24\textwidth]{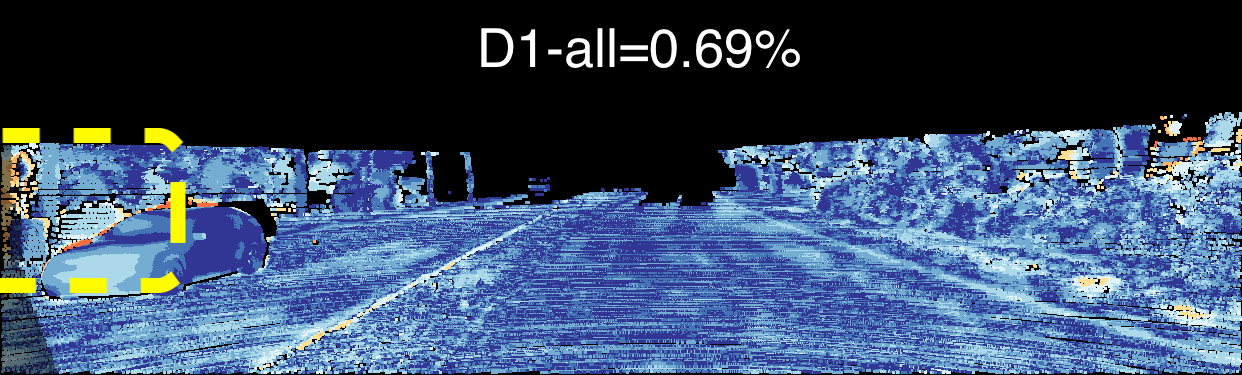} &
        \includegraphics[width=0.24\textwidth]{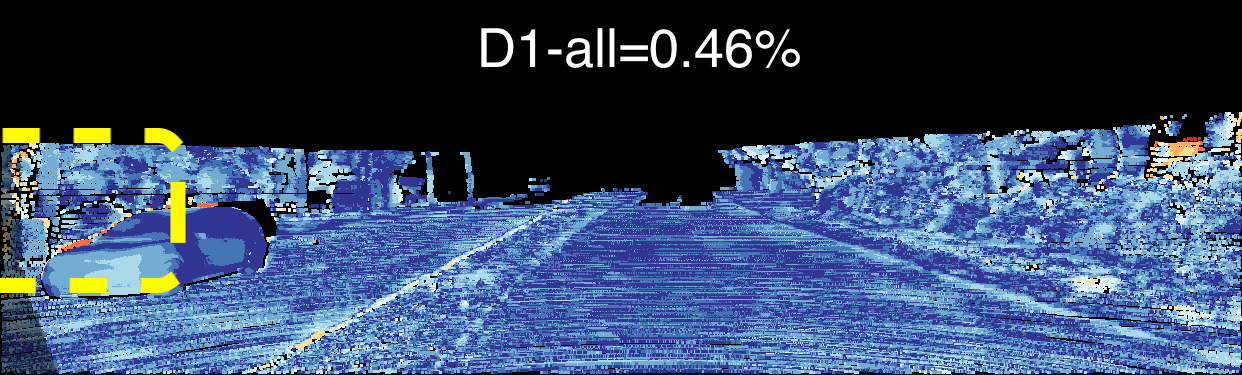} \\
        \midrule
        \includegraphics[width=0.24\textwidth]{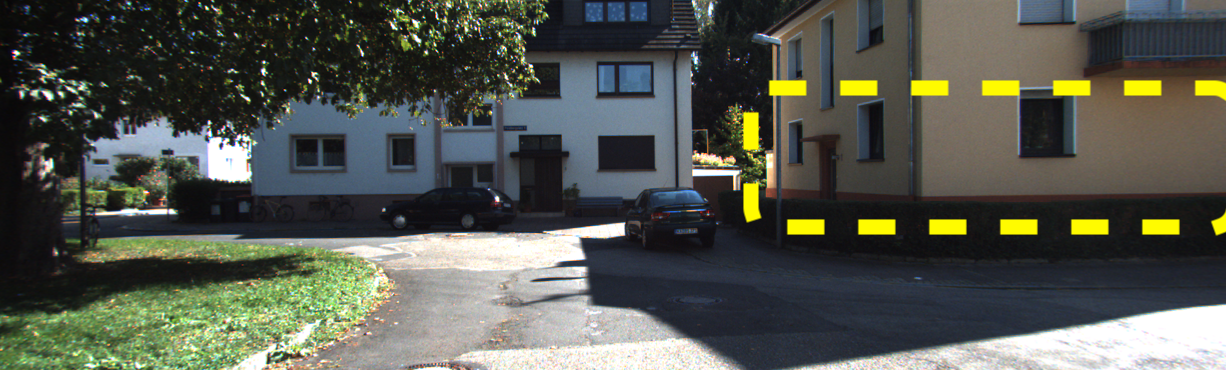} &
        \includegraphics[width=0.24\textwidth]{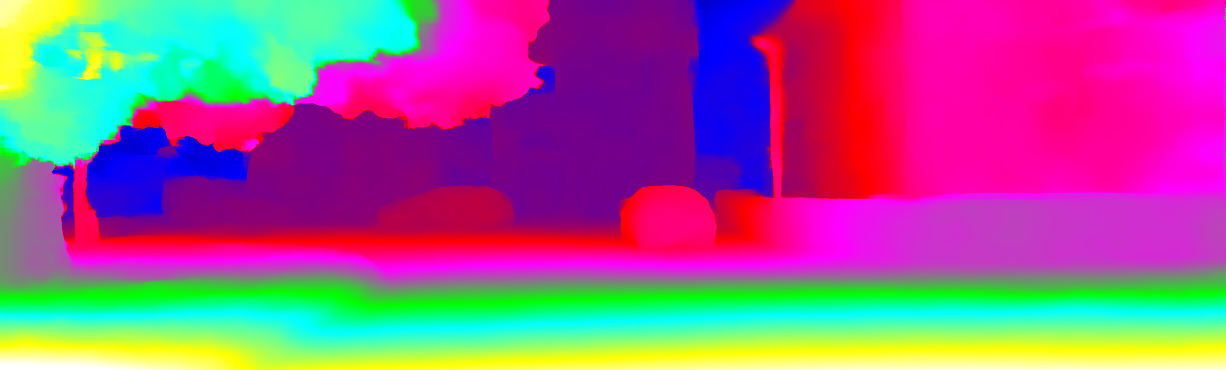} &
        \includegraphics[width=0.24\textwidth]{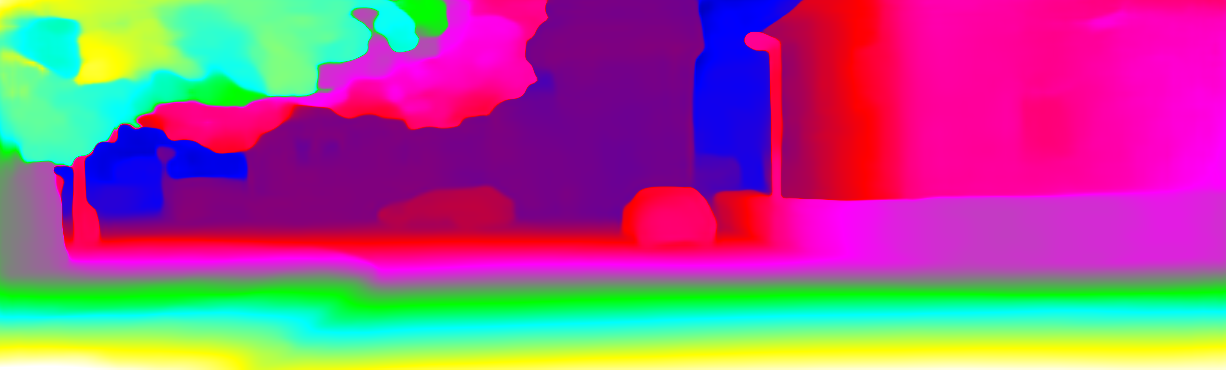} &
        \includegraphics[width=0.24\textwidth]{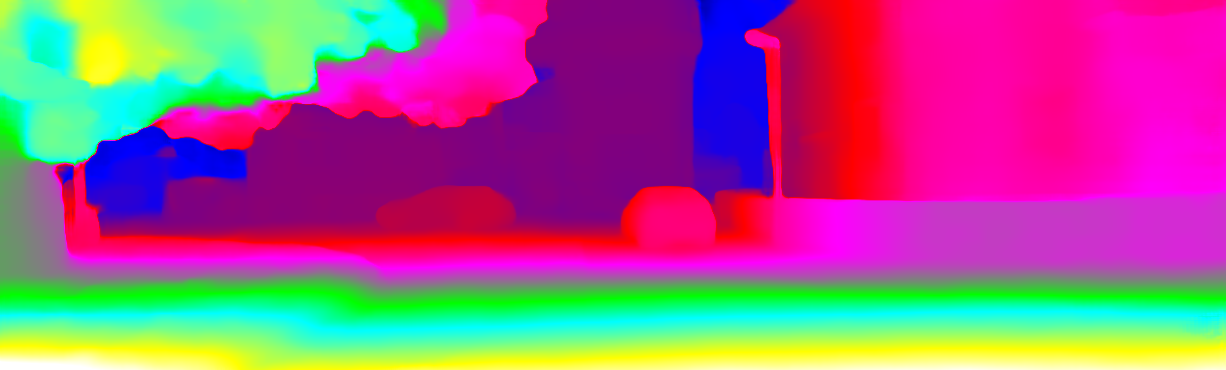} &
        \includegraphics[width=0.24\textwidth]{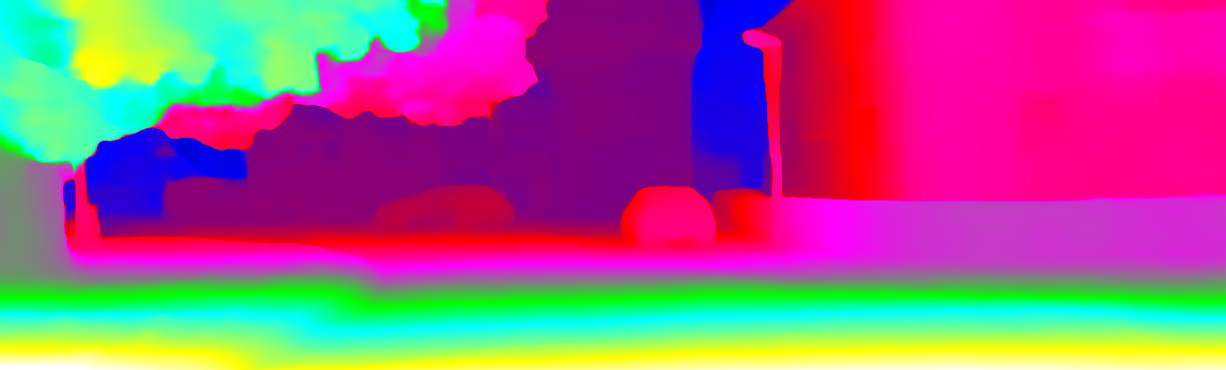} \\
        \includegraphics[width=0.24\textwidth]{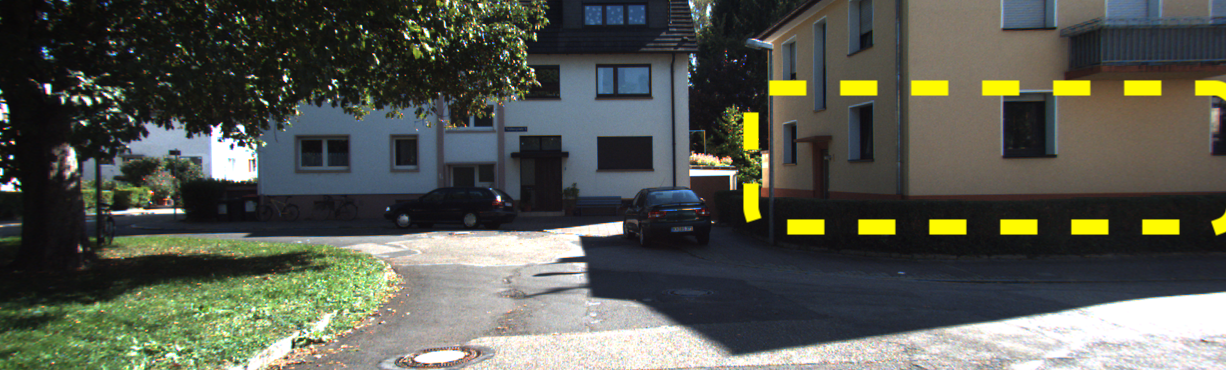} &
        \includegraphics[width=0.24\textwidth]{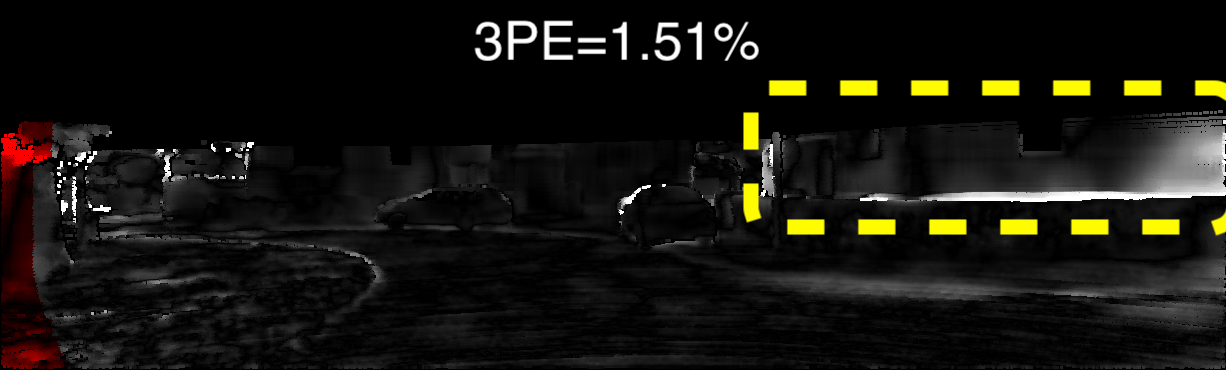} &
        \includegraphics[width=0.24\textwidth]{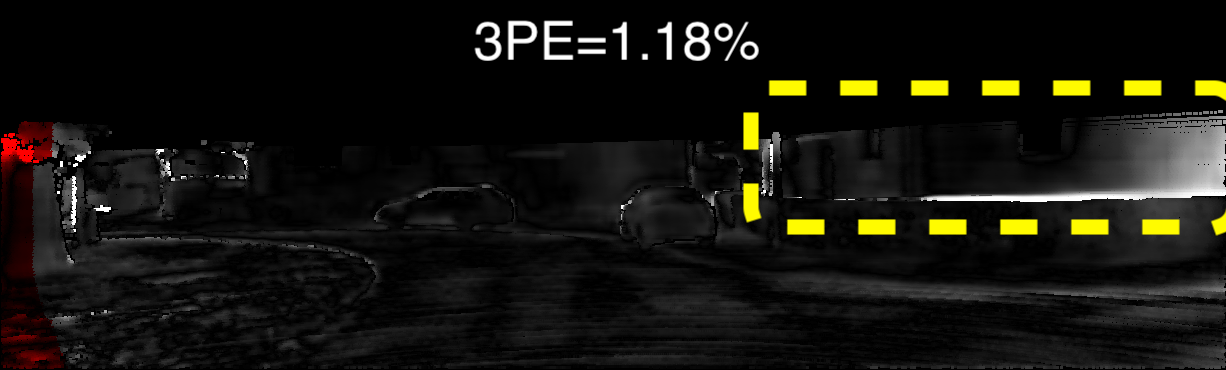} &
        \includegraphics[width=0.24\textwidth]{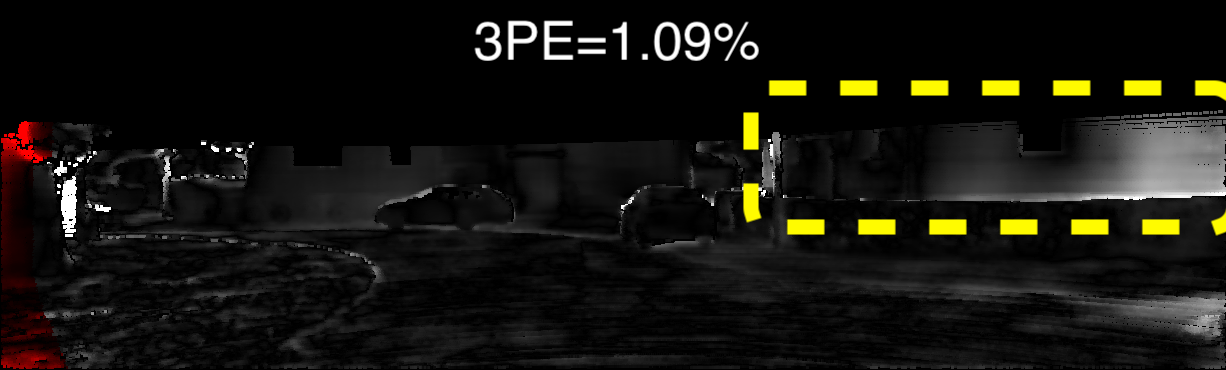} &
        \includegraphics[width=0.24\textwidth]{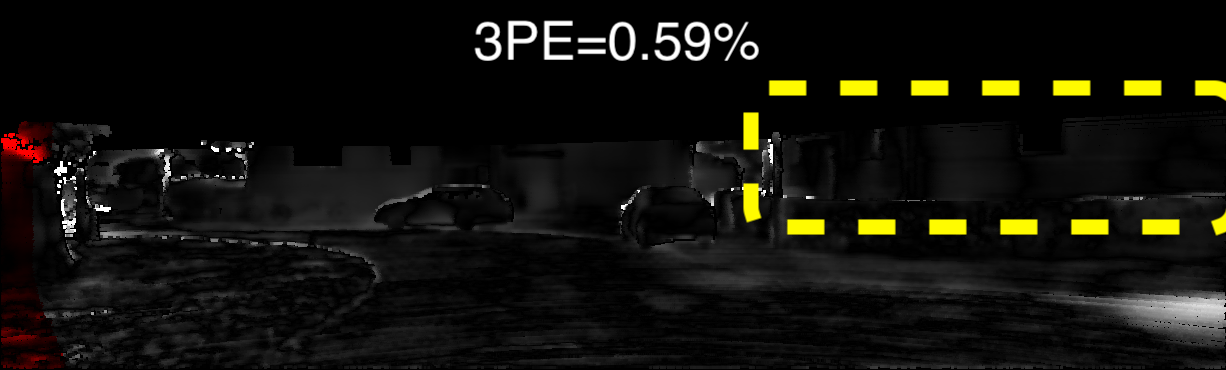} \\
    \end{tabular}
    }
    \vspace{-0.2cm}
    \caption{\textbf{Results on KITTI 2015 and 2012 testing set.} Compared to existing methods \cite{coex,LEAStereo}, \net{} exploits time to improve accuracy at occlusions (top) or texturless regions (bottom). For the error maps on second and forth row, colder and darker color means lower the error respectively.
    }
    \label{fig:kitti_benchmark}
\end{figure*}

\section{Video Demo}
For a better understanding of the effect of \net{} both in single-pair and temporal mode when dealing with a continuous video, and also as a comparison with state-of-the-art method~\cite{coex}, we conduct inference on another KITTI raw sequence '2011\_09\_28\_drive\_0037' (a campus scene, its poses are computed by ORB-SLAM3~\cite{ORBSLAM3} algorithm.) which doesn't appear in any our training process. The results are displayed in the appended \textbf{video demo}. As shown in the video, even without access to the ground truth pose, our \net{} in temporal mode is robust to dynamic objects (\eg{}, pedestrians). Benefiting from past context, for both static and dynamic regions, it provides more impressive predictions compared to CoEx~\cite{coex} and our model in single-pair mode, e.g., accurate estimations on occluded regions and fine-grained details on pillars of the building and street signs.

\section{Implementation Details}

We implement our network in PyTorch~\cite{pytorch}, using RMSProp as optimizer to train all models in an end-to-end fashion. We train the models from scratch for 40, 40 epochs on SceneFlow and TartanAir respectively. Given the pretrained model from SceneFlow, TartanAir or KITTI Raw Sequences~\cite{eigen2014depth} as we discussed in Section 4.2 and Tab.5 of the main manuscript, we finetune our model for 16, 16 epochs on KITTI 2012 and KITTI 2015 separately. Scale weights are $\lambda_0=1.0$, $\lambda_1=0.5$, $\lambda_2=0.7$, $\lambda_3=2.0$, $\lambda_{final}=2.0$, while we set $K=2$ for top-$K$ selection, number of candidates $n=12$ in stage 1 while $n=5$ in stages $s=2,3$, $\beta=4$ for sampling and $N_{key}=3$ in Local Map. As for keyframe selection strategy, we set $|\textbf{t}_{max}|= 0.1$m and $|\textbf{R}_{max}| = 15^{\circ}$ respectively. We leverage EfficientNetV2-S~\cite{tan2021efficientnetv2} as backbone feature extractor, while the hourglass network proposed in~\cite{GWCNet} for cost aggregation. For all 3D convolutions with kernel size $k\times k\times k$ where $k>1$, we implement it in depthwise~\cite{mobilenetv2} manner which consists two convolutions with kernels of size $k\times1\times1$ and $1\times k \times k$ respectively to save computation. For all datasets, we perform asymmetric chromatic augmentation as described in~\cite{HSM} to mitigate the effect of varying lighting and exposure. Furthermore, we also add random patching~\cite{HSM} on the right image to help the network deal with occluded areas. The maximum disparity value is $D_{max}=192$. The smooth $L_1$ loss, \ie{}, Huber loss used in Eq.(4) of the main manuscript is defined as:

\begin{equation}
    \begin{aligned}
    smooth_{L_{1}}(x) = 
    \Big\{
    \begin{array}{lc}
          0.5x^{2}, & \textrm{if}\; |x|<1,  \\
          |x| - 0.5, & \textrm{otherwise}.
    \end{array}
    \end{aligned}
\end{equation}

Furthermore, \textbf{our code} will be publicly available after the paper is accepted.

\section{Dataset Description and Qualitative Results}

\begin{figure*}[t]
	\centering
		\includegraphics[width=2.0\columnwidth]{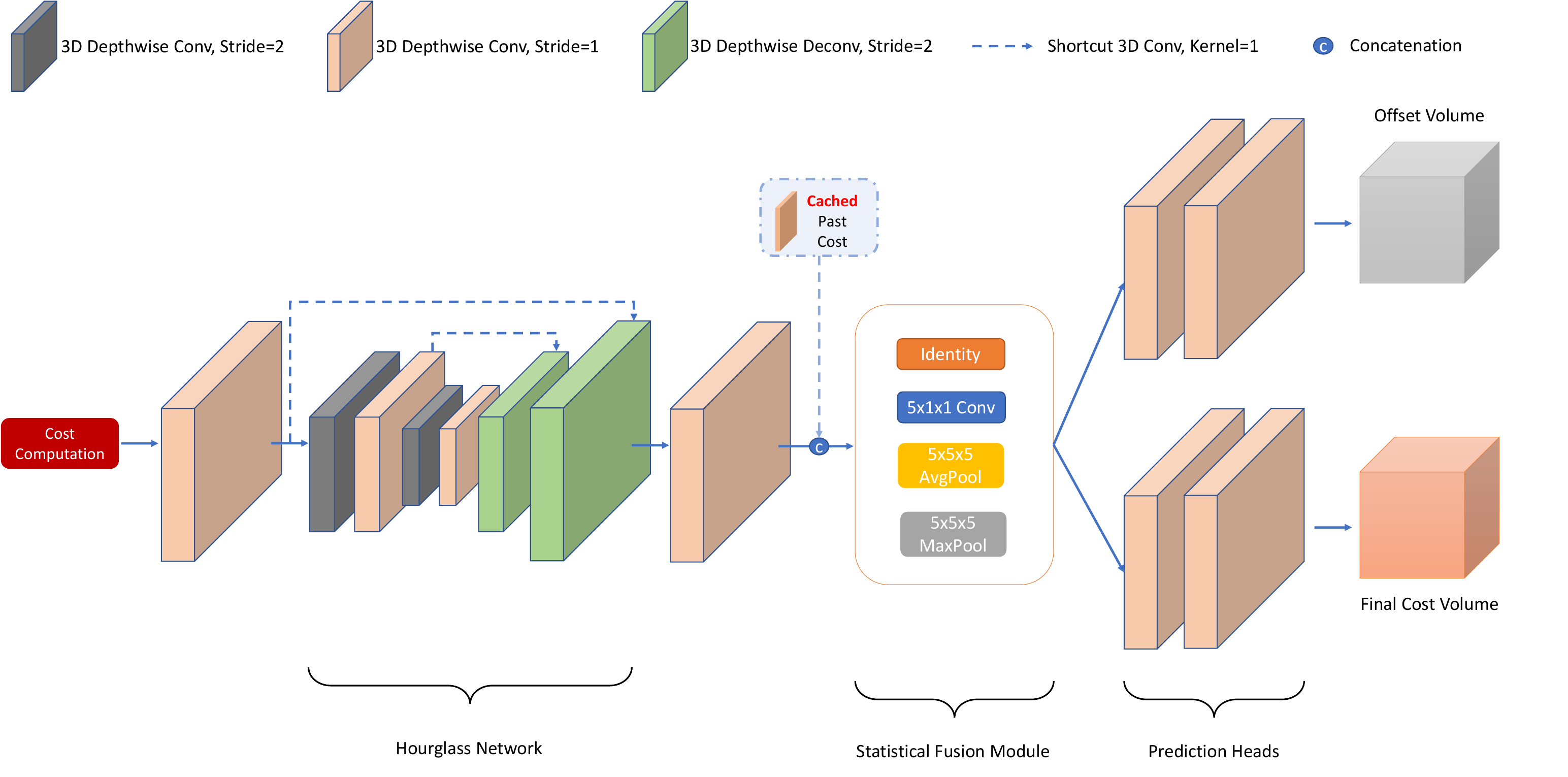}
		\caption{\textbf{The structure of our proposed network at stage 2.} 
		} 
	\label{fig:stage2}
\end{figure*}

\subsection{SceneFlow}
It is a large synthetic dataset~\cite{sceneflow} including 35,454 training and 4,370 test images with a resolution of $540\times960$. We only use the Flyingthings part in "finalpass" format for training and testing. Specifically, we randomly crop the image of $512\times960$ and set the batch size as 4. Since it does not provide ground truth camera poses, we use this dataset for single-pair mode only. The model is trained for 40 epochs with the initial learning rate of 0.001 decaying by a factor of 0.1 at epochs 30.

We validate the effectiveness of our model in single-pair mode on SceneFlow dataset and qualitative results are shown in
Fig. \ref{fig:qualitative-sceneflow}. We can notice that our network achieves extremely low error in both cases.

\begin{table}[h]
	\begin{center}
        \resizebox{0.4\columnwidth}{!}{
			\begin{tabular}{c|c}
			    \toprule
				Scene & Part \\
                \midrule
                abandonedfactory & P002 \\
                amusement & P007 \\
                carwelding & P003 \\
                endofworld & P006 \\
                gascola & P001 \\
                hospital & P042 \\
                office & P006 \\
                office2 & P004 \\
                oldtown & P006 \\
                seasonforest & P002 \\
                seasonforest\_winter & P015 \\
                soulcity & P008 \\
			    \bottomrule	
			\end{tabular}
		}
	\end{center}
	\caption{\textbf{TartanAir Testing Dataset.} We list Scene-Part pairs used in our testing set.}
	\label{tab:tartanair-testing}
\end{table}

\subsection{TartanAir} 
It's a challenging synthetic dataset~\cite{TartanAir} with moving objects and various light and weather conditions. To adapt it for stereo matching, we manually split the dataset with hard motion (which has 6DoF motion, the max translation and rotation are 0.5 meters and $10^{\circ}$ respectively.) for training and testing. Specifically, the scenes consisted the testing dataset are listed in ~\cref{tab:tartanair-testing}. The other parts of the corresponding scene are used for training. Overall, we collect nearly 66K stereo pairs for training and 5K for testing. The model is trained for 40 epochs with the initial learning rate of 0.001, reduced to 0.0001 at epoch 30. As for the first 20 epochs, we train the model in the single-pair mode to let the network learn to perform stereo matching. Then, we set the model in the temporal mode for the left 20 epochs. For all experiments, we keep the image at full resolution (i.e., $480\times640$) with batch size 16, and use the same training schedule for all experiments on the TartanAir dataset. As the TartanAir dataset targets at visual SLAM and thus ground truth poses are provided. In temporal experiments, we leverage ground truth poses for warping.

\subsection{KITTI Raw Sequences}
As we discussed in the main paper, we augment the KITTI~\cite{KITTI2012,KITTI2015} datasets with KITTI Raw Sequences~\cite{eigen2014depth}. Specifically, the KITTI Raw Sequences are composed of several outdoor scenes captured with car-mounted cameras and depth sensors. Following~\cite{eigen2014depth}, we get 61 stereo video sequences (containing 42K pairs) with pseudo labels for pretraining, and poses are calculated from the GPS/OXTS devices on KITTI. As for pseudo label generation, we leverage the prediction results from LEAStereo~\cite{LEAStereo} which has been finetuned on KITTI 2015~\cite{KITTI2015} training dataset. We also perform left-right consistency~\cite{lrconsistency} to filter out outliers in the generated pseudo labels. Given the model trained on SceneFlow or TartanAir dataset, we further pretrain the network for 10 epochs with an initial learning rate of 0.001 and decrease it to 0.0001 at epochs 8.
Besides, we set the batch size as 4 and randomly crop the image into $320\times1024$. 

\subsection{KITTI 2012\&2015} 

They are both real-world datasets collected by a driving car and depicting urban scenes. KITTI 2015~\cite{KITTI2015} contains 200 training and 200 testing stereo pairs while KITTI 2012~\cite{KITTI2012} has 194 and 195 pairs for training and test, respectively. Both the splits provide sparse ground truth depth labels. For temporal mode evaluation, we leverage the multi-view split, which contains, for each sample, also the 10 previous pairs. We estimate camera poses with an off-the-shelf SLAM algorithm~\cite{ORBSLAM3}. We randomly crop the image into resolution $320\times1024$, use batch size as 4, and train our models on KITTI 2012~\cite{KITTI2012} and KITTI 2015~\cite{KITTI2015} with the same schedule. More specifically, for both single-pair and temporal mode, we use the pre-trained model from KITTI Raw Sequences~\cite{eigen2014depth} and finetune it for 16 epochs with the initial learning rate of 0.0001 and reduce it by a factor of 0.1 at epochs 12. 

Fig. \ref{fig:kitti_benchmark} and appended \textbf{video demo} show qualitative comparisons between existing networks~\cite{coex,LEAStereo} and \net{}, highlighting the benefits yielded by temporal mode.

\section{Model Architecture Details}
Tab.~\ref{network parameters} presents the details of the TemporalStereo which is used in experiments to produce state-of-the-art accuracy on Scene Flow dataset~\cite{sceneflow}, TartanAir~\cite{TartanAir} and KITTI benchmarks~\cite{KITTI2012,KITTI2015}. The feature extraction backbone is based on EfficientNetV2-S~\cite{tan2021efficientnetv2} and further equipped with Residual Temporal Shift Module (TSM)~\cite{tsm} to encode temporal info. For a better understanding of our architecture, we give an illustration of Stage 2 in Fig~\ref{fig:stage2}.

\begin{table*}[!ht]
	\begin{center}
        \resizebox{1.8\columnwidth}{!}{
			\begin{tabular}{c|c|c}
				\toprule
				 Name & Layer setting & Output dimension \\

				 \midrule
				 
				 \multicolumn{3}{c}{\textbf{Feature Extraction}} \\
				 
				 \midrule
				 
				 input & & $H \times W \times 3$ \\
				 \hline
				 conv\_stem & EfficientNetV2-S.Conv$3\times3$ & $\frac{1}{2}H\times \frac{1}{2}W \times 24$ \\
				 \hline
				 block0 & EfficientNetV2-S.Stage1 & $\frac{1}{2}H\times \frac{1}{2}W \times 24$ \\
				 \hline
				 block1 & EfficientNetV2-S.Stage2 & $\frac{1}{4}H\times \frac{1}{4}W \times 48$ \\
				 \hline
				 block2 &  EfficientNetV2-S.Stage3 & $\frac{1}{8}H\times \frac{1}{8}W \times 64$ \\
				 \hline
				 block3 & $ \begin{array}{c} \textrm{ EfficientNetV2-Stage4,5} \\ \textrm{with Residual TSM~\cite{tsm} at each Inverted Residual Block} \end{array} $ & $\frac{1}{16}H\times \frac{1}{16}W \times 160$ \\
				 \hline
				 block4 &$ \begin{array}{c} \textrm{ EfficientNetV2-S.Stage6,7} \\ \textrm{with Residual TSM~\cite{tsm} at each Inverted Residual Block} \end{array} $ & $\frac{1}{32}H\times \frac{1}{32}W \times 272$ \\
				 \hline
				 conv\_out4 & $ \left[ 3\times 3, 320 \right] $ & $\frac{1}{32}H\times \frac{1}{32}W \times 320$ \\
				 \hline
				 conv\_out3 & $\left[ \begin{array}{c} \textrm{concat [conv\_out4}\uparrow \textrm{, block3]} \\ 3\times 3, 256 \\ 3\times 3, 256  \end{array} \right] $ & $\frac{1}{16}H\times \frac{1}{16}W \times 256$ \\
				 \hline
				 conv\_out2 & $\left[ \begin{array}{c} \textrm{concat [conv\_out3}\uparrow \textrm{, block2]} \\  3\times 3, 128 \\ 3\times 3, 128  \end{array} \right] $ & $\frac{1}{8}H\times \frac{1}{8}W \times 128$ \\
				 \hline
				 conv\_out1 & $\left[ \begin{array}{c} \textrm{concat [conv\_out2}\uparrow \textrm{, block1]} \\  3\times 3, 64 \\ 3\times 3, 64  \end{array} \right] $ & $\frac{1}{4}H\times \frac{1}{4}W \times 64$ \\
				 
				 \midrule
				 
				 \multicolumn{3}{c}{\textbf{Stage 1}} \\
				 
				 \midrule
				 
				 cost\_volume & $ \begin{array}{c} \textrm{concat  and multiscale groupwise} \\ \textrm{left and shifted right conv\_out3} \end{array} $ & $ \frac{1}{16}D\times \frac{1}{16}H\times \frac{1}{16}W \times 704$ \\
				 \hline
				 
				 aggregation\_1 & $\left[ \begin{array}{c} 1\times 3\times 3, 32, \quad 3\times 1\times 1, 32 \\ \textrm{Hourglass Network~\cite{GWCNet}} \\ 1\times 3\times 3, 32, \quad 3\times 1\times 1, 32   \end{array} \right] $ & $\frac{1}{16}D\times \frac{1}{16}H\times \frac{1}{16}W \times 32$ \\ 
				 \hline
				 fusion\_1 & $\left[ \begin{array}{c} \textrm{concat PastCost / Identity} \\ \textrm{Statistical Fusion Module}  \end{array} \right] $   & $(\frac{1}{16}D + \textrm{top-}K)\times \frac{1}{16}H\times \frac{1}{16}W \times 32$ \\ 
				 \hline
				 output\_1 & $\left[ \begin{array}{c} \textrm{Prediction Heads} \\ \textrm{Convex Upsample}  \end{array} \right] $  & $\frac{1}{8}H\times \frac{1}{8}W $ \\ 
				 
				 \midrule
				 
				 \multicolumn{3}{c}{\textbf{Stage 2}} \\
				 
				 \midrule
				 
				 cost\_volume & $ \begin{array}{c} \textrm{concat  and multiscale groupwise} \\ \textrm{left and shifted right conv\_out2} \end{array} $ & $ (n + N_{key})\times \frac{1}{8}H\times \frac{1}{8}W \times 294$ \\
				 \hline
				 
				  aggregation\_2 & $\left[ \begin{array}{c} 1\times 3\times 3, 16, \quad 3\times 1\times 1, 16 \\ \textrm{Hourglass Network~\cite{GWCNet}} \\ 1\times 3\times 3, 16, \quad 3\times 1\times 1, 16   \end{array} \right] $ & $(n+N_{key})\times \frac{1}{8}H\times \frac{1}{8}W \times 16$ \\ 
				 \hline
				 fusion\_2 & $\left[ \begin{array}{c} \textrm{concat PastCost / Identity} \\ \textrm{Statistical Fusion Module}  \end{array} \right] $   & $(n+N_{key} + \textrm{top-}K)\times \frac{1}{8}H\times \frac{1}{8}W \times 16$ \\ 
				 \hline
				 output\_2 & $\left[ \begin{array}{c} \textrm{Prediction Heads} \\ \textrm{Convex Upsample}  \end{array} \right] $  & $\frac{1}{4}H\times \frac{1}{4}W $ \\ 
				 \midrule \multicolumn{3}{c}{\textbf{Stage 3}} \\
				 \midrule
				 cost\_volume & $ \begin{array}{c} \textrm{concat  and multiscale groupwise} \\ \textrm{left and shifted right conv\_out1} \end{array} $ & $ n \times \frac{1}{4}H\times \frac{1}{4}W \times 156$ \\
				 \hline
				 aggregation\_3 & $\left[ \begin{array}{c} 1\times 3\times 3, 8, \quad 3\times 1\times 1, 8 \\ \textrm{Hourglass Network~\cite{GWCNet}} \\ 1\times 3\times 3, 8, \quad 3\times 1\times 1, 8   \end{array} \right] $ & $n\times \frac{1}{4}H\times \frac{1}{4}W \times 8$ \\ 
				 \hline
				 output\_3 & $\left[ \begin{array}{c} \textrm{Prediction Heads} \\ \textrm{Upsample Module from~\cite{coex}}  \end{array} \right]$ & $H\times W$ \\ 
				 \bottomrule	
			\end{tabular}
		}
	\end{center}
	\caption{Parameters of the network architecture of TemporalStereo. $\uparrow$ means bilinearly upsampling with a factor of 2. $n$ is the number of disparity candidates when applying inverse transform sampling, $N_{key}$ is the number of last keyframes in the memory bank, while top-$K$ is the number of selected values in aggregated cost volume.}
	\label{network parameters}
\end{table*}

\end{document}